\documentclass[letterpaper]{article} 
\usepackage{aaai25}  
\usepackage{times}  
\usepackage{helvet}  
\usepackage{courier}  
\usepackage[hyphens]{url}  
\usepackage{graphicx} 
\usepackage{natbib}  
\usepackage{caption} 
\usepackage{algorithm}
\usepackage{algorithmic}

\usepackage{newfloat}
\usepackage{listings}
\DeclareCaptionStyle{ruled}{labelfont=normalfont,labelsep=colon,strut=off} 
\floatstyle{ruled}
\newfloat{listing}{tb}{lst}{}
\floatname{listing}{Listing}
\usepackage{subcaption}
\usepackage{amsmath}
\usepackage{multirow}
\usepackage{booktabs}
\usepackage[pagebackref,colorlinks]{hyperref}
\usepackage{xcolor}
\usepackage{appendix}

\newcommand{\LEAST}{GALA}
\newcommand{\good}{good}  
\newcommand{\bad}{bad}  

\title{A Layer Selection Approach to Test Time Adaptation}
\author {
    Sabyasachi Sahoo\textsuperscript{\rm 1,2},
    Mostafa ElAraby\textsuperscript{\rm 2,3},
    Jonas Ngnawe\textsuperscript{\rm 1,2},
    Yann Batiste Pequignot\textsuperscript{\rm 1},\\
    Frédéric Precioso\textsuperscript{\rm 4},
    Christian Gagné\textsuperscript{\rm 1,2,5}
}
\affiliations {
    \textsuperscript{\rm 1}IID, Université Laval\\
    \textsuperscript{\rm 2}Mila\\
    \textsuperscript{\rm 3}Université de Montréal\\
    \textsuperscript{\rm 4}Université Cote d'Azur, CNRS, INRIA, I3S, Maasai\\
    \textsuperscript{\rm 5}Canada CIFAR AI Chair\\
    sabyasachi.sahoo.1@ulaval.ca
}

\begin{document}

\maketitle

\begin{abstract}
Test Time Adaptation (TTA) addresses the problem of distribution shift by adapting a pretrained model to a new domain during inference. When faced with challenging shifts, most methods collapse and perform worse than the original pretrained model. In this paper, we find that not all layers are equally receptive to the adaptation, and the layers with the most misaligned gradients often cause performance degradation. To address this, we propose GALA, a novel layer selection criterion to identify the most beneficial updates to perform during test time adaptation. This criterion can also filter out unreliable samples with noisy gradients. Its simplicity allows seamless integration with existing TTA loss functions, thereby preventing degradation and focusing adaptation on the most trainable layers. This approach also helps to regularize adaptation to preserve the pretrained features, which are crucial for handling unseen domains. Through extensive experiments, we demonstrate that the proposed layer selection framework improves the performance of existing TTA approaches across multiple datasets, domain shifts, model architectures, and TTA losses.

\end{abstract}

\section{Introduction}

\begin{figure}[t]
    \centering
    \begin{subfigure}{0.4\linewidth}
        \includegraphics[width=\linewidth]{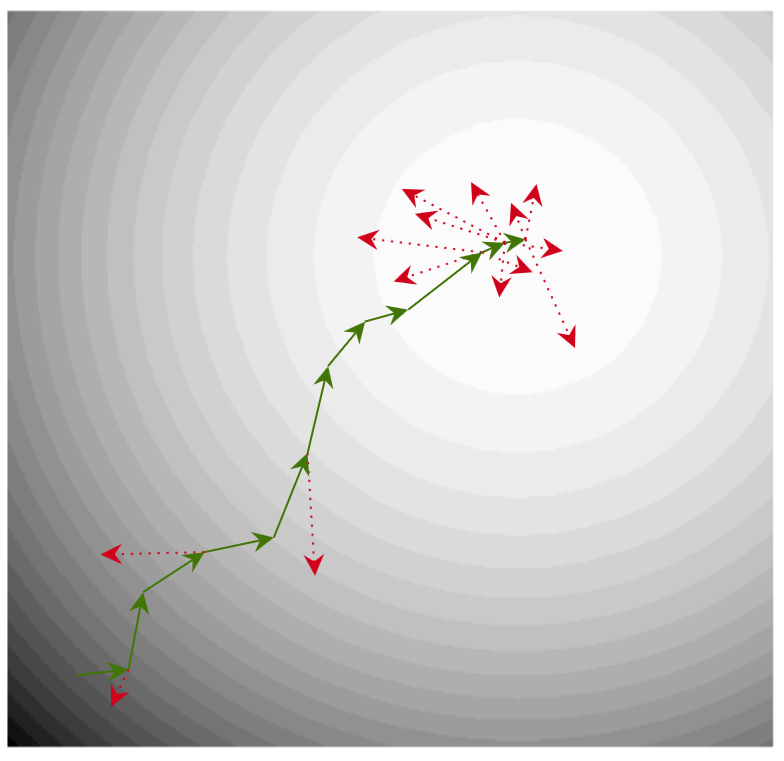}
        \centering
        \caption{Gradient alignment}
    \end{subfigure}
    \hspace{0.3cm}
    \begin{subfigure}{0.49\linewidth}
        \includegraphics[width=\linewidth]{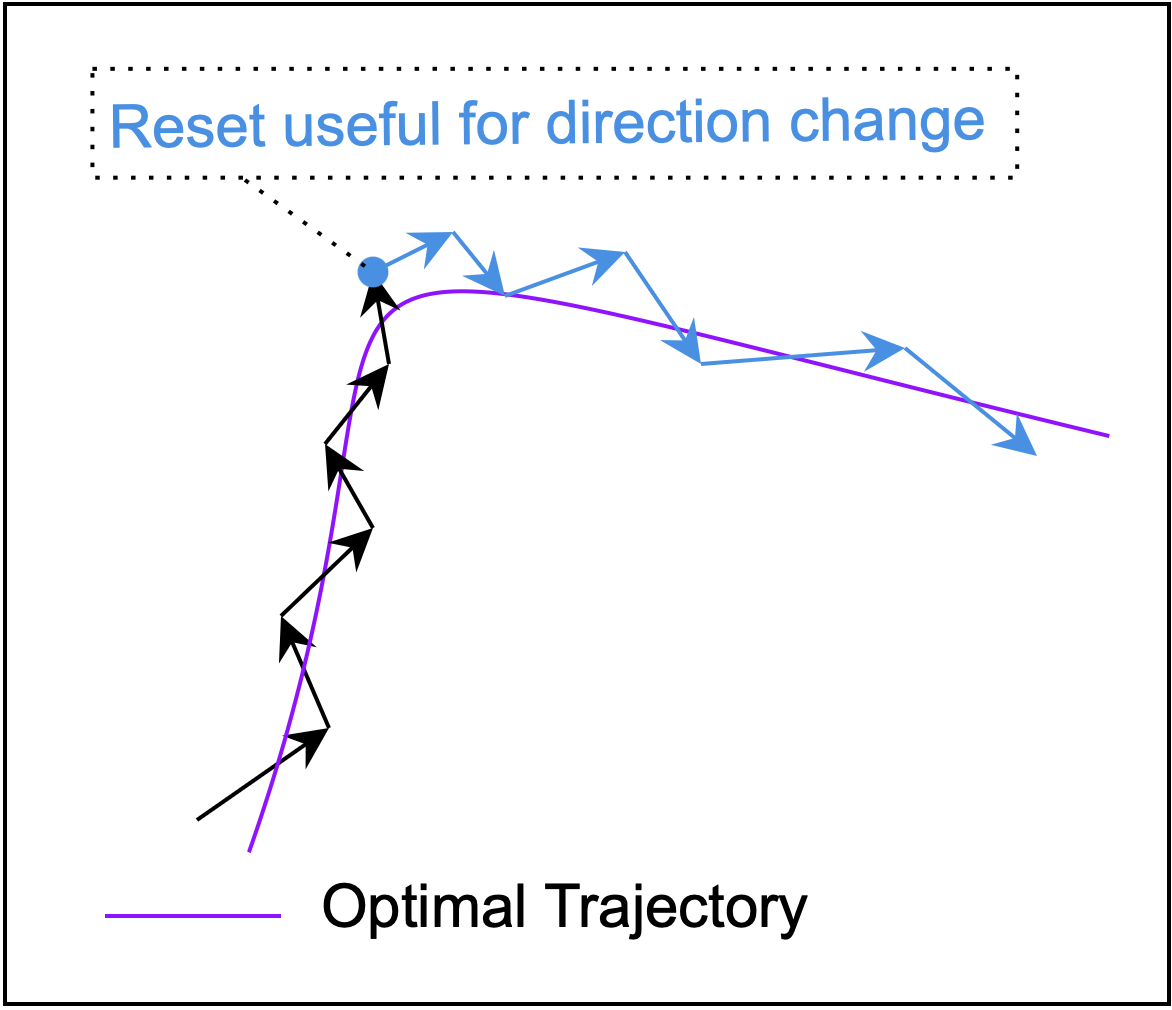}
        \centering
        \caption{Helpful reset}
    \end{subfigure}
    \caption{Intuition for proposed approaches: \textbf{(a)} As the model reaches closer to minima, the individual sample gradients start to be misaligned with gradients of previous samples~\cite{mahsereci2017early, forouzesh2021disparity, agarwal2022estimating}. We leverage this misalignment to identify trainable layers. \textbf{(b)} While effective in moving in the direction of most aligned gradients, the introduced criterion based on angular deviation could prevent adaptation when a direction change is needed, even if the following updates (or gradients) are aligned. A reset of the past horizon (i.e., gradients of previous samples) considered in the alignment condition can help resolve such situations.
    }
    \label{fig:cosine-intuition}
\end{figure}

Distribution shifts \cite{gulrajani2021search} present significant challenges when deploying deep learning models in real-world scenarios. Test Time Adaptation (TTA) \cite{liang2023comprehensive} has emerged as a promising approach for adapting pretrained models to novel domains during inference. However, these methods often falter when confronted with severe or diverse distributional changes. To mitigate potential performance degradation, various regularization strategies have been proposed \cite{niu2022efficient, shin2024gradient}. Nevertheless, these strategies might not effectively address all types of shifts or TTA losses \cite{burns2021limitations, zhao2023pitfalls}. Moreover, the selection of layers in the existing TTA approaches typically remains unchanged across different shifts \cite{wang2024search}, which may not be optimal. In contrast, layer selection has demonstrated substantial improvements in related fields such as domain generalization \cite{2020EccvDMG}, fine-tuning \cite{lee2023surgical}, multi-task learning \cite{wallingford2022task}, and continual learning \cite{zhao2023does}, underscoring the importance and broad potential of layer selection. Still, the question of optimal layer selection remains largely unexplored in the context of TTA.

\begin{figure*}[t]
    \centering
    \includegraphics[width=0.7\linewidth]{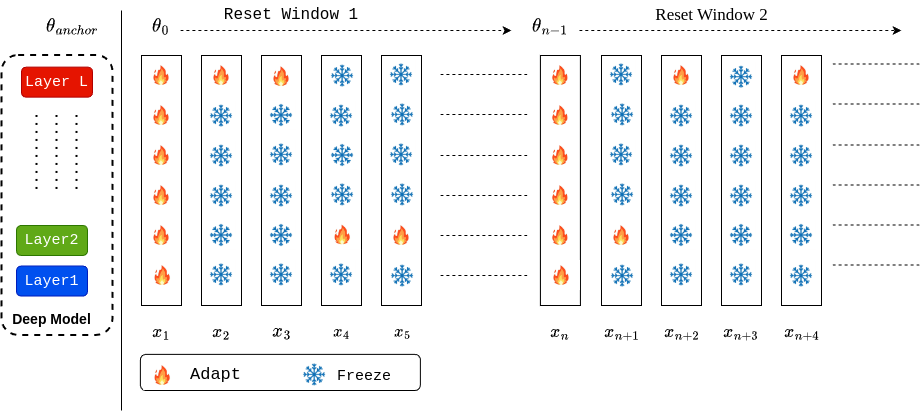}
    \caption{Gradient-Aligned Layer Adaptation or \LEAST{} framework adapts the most gradient-aligned layer per sample. It adapts all the layers for the first sample in a reset window (e.g., $x_1, x_n, \dots$). For all the other samples, it adapts the most gradient-aligned layer per sample. It can also skip the adaptation on a given sample if all the layers are misaligned. We use a reset window to periodically reset the anchor parameters to allow for a change in direction.}
    \label{fig:mask}
\end{figure*}

In this paper, we study layer selection for TTA  and show that not all layers of a given model are equally receptive to adaptation. Our findings suggest that adapting the right layer can lead to meaningful improvement, while adapting the wrong layer can cause significant performance degradation in TTA approaches. Specifically, we find that while adapting a certain layer may benefit one shift, it may be detrimental to another. Additionally, we find that on a given shift, the effect of adapting a certain layer also depends on the loss used. Therefore, while we observe an important potential in selecting the right layer to adapt in each situation, identifying these layers at test time can be challenging.

To address the challenges of layer selection, we propose Gradient-Aligned Layer Adaptation, \LEAST{}, a novel criterion to identify good layers for adaptation at test time. \LEAST{} ranks all the layers of a model based on the gradient alignment of the current adaptation step. As the model approaches the optimization minima, the variance in gradient updates increases \cite{mahsereci2017early, forouzesh2021disparity, agarwal2022estimating}, leading to potential overfitting and performance degradation.
Building on this insight, for each layer, we propose to measure the angle deviation of the proposed gradient update from the average of all gradient updates performed so far (including the proposed one). This measure can also be expressed as the cosine between the proposed update and the (anticipated) total displacement of the parameters from their pretrained values. This allows us to compare the updates for each layer on a common scale and only perform the update of the layer with the smallest angle.

Our extensive experiments on Domainbed \cite{gulrajani2021search} and Continual TTA benchmark \cite{wang2022continual} demonstrate that \LEAST{} consistently surpasses \emph{all layers} and \emph{ERM} (no adaptation) baselines and other existing layer selection baselines across various datasets, various neural network backbones, and various losses. Further analysis reveals that \LEAST{} can identify the \good{} layers, which exhibit significant displacement in a single direction and higher gradient alignment. This layer selection strategy enhances the model's ability to adapt to novel domains by mitigating performance degradation and potentially serves as a regularization mechanism, reducing catastrophic forgetting of source domain knowledge. Ablation studies reveal that \LEAST{}’s performance is robust to hyperparameter choices.

The contributions of our paper are summarized as follows:
\begin{enumerate}
    \item We study the problem of layer selection for TTA and find that while adapting specific layers can enhance performance, the optimal set of layers for adaptation is not universal but rather contingent upon the particular distribution shift encountered and the TTA loss function employed during inference.
    \item We introduce \LEAST{}, a novel layer selection criterion to identify good layers to adapt per sample that can be applied across various distribution shifts and TTA loss functions at test time.
    \item Through extensive experiments across different backbones, datasets, and TTA losses, we show that \LEAST{} outperforms standard \emph{ERM} (no adaptation), \emph{all layers} baselines, and other layer selection baselines (i.e., AutoRGN and AutoSNR \cite{lee2023surgical}) for TTA.
\end{enumerate}

\section{Proposed Approach}
In the following, we describe the Gradient-Aligned Layer Adaptation (\LEAST{}) framework for Test Time Adaptation (TTA). We first introduce our layer selection framework for TTA (Sec.\ \ref{subsec:LEASTframework}), before describing the cosine distance criterion proposed to identify the most trainable layers (Sec.\ \ref{subsec:LEASTcosine}), and then present the reset window strategy used to improve performances with the proposed cosine criterion (Sec.\ \ref{subsec:LEASTresetWindow}).

\subsection{Layer selection framework for TTA}
\label{subsec:LEASTframework}

\begin{figure*}[t]
  \centering
    \includegraphics[width=0.7\linewidth]{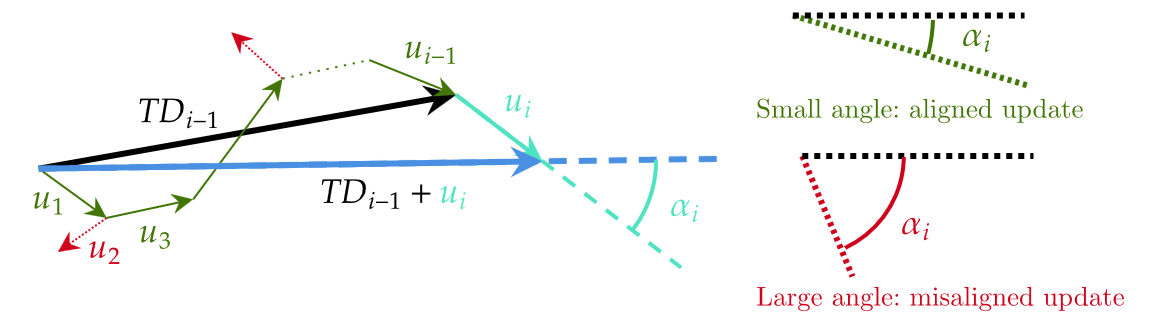}
  \caption{Illustration of proposed criterion based on angular deviation. Different layers can be ranked based on their alignments with previous gradient updates. In the figure, updates drawn in red are discarded, while green updates are applied, adding up to $\mathbf{TD}_{i-1}$. The update under scrutiny $\mathbf{u_i}$ is drawn in cyan, and its sum with $\mathbf{TD}_{i-1}$ is drawn in blue. Application of update $\mathbf{u_i}$ or not is based on the angle $\alpha_i$.}
  \label{fig:methodology}
\end{figure*}

Let $f_{\mathbf{\theta}_\mathrm{src}}$ denote the model parameterized by parameters $\mathbf{\theta}_\mathrm{src}$ trained beforehand on the source domain $\mathcal{D}_\mathrm{src}$. Let us also assume that target domain samples $\{x_i\}_{i=1}^n$ are coming in an online fashion at test time. For some sample $x_i$ at test time, TTA adapts the model to obtain $\theta_i$ before performing inference \cite{sun2020test, liang2023comprehensive}. We set $\mathbf{\theta}_0=\mathbf{\theta}_\mathrm{src}$ and, at each step, $\theta_i$ is obtained by updating $\theta_{i-1}$ using the following equation:
\begin{equation}    \label{eq:tta}
\mathbf{\theta}_{i} = \mathbf{\theta}_{i-1} + \mathbf{u}_i, 
\end{equation}
where $\mathbf{u}_i$ is a parameter update specific to the TTA algorithm. Typically, if SGD optimizer is used with learning rate $\eta$, this update takes the form $\mathbf{u}_i=-\eta \nabla \mathcal{L}(x_i; \mathbf{\theta}_{i-1})$, where $\mathcal{L}$ is the unsupervised loss specific to the TTA method. 

In this section, we consider single-step TTA performed online on a single input sample using an SGD optimizer for notation simplicity.
Throughout, we assume the deep learning model is written as a certain composition of functions, which we simply refer to as layers, though any granularity would do.
This allows us to write the model at step $i$ as $f_{\theta_i} = f_{\theta_{i,L}} \circ \dots \circ f_{\theta_{i,1}}$, where $\theta_{i,l}$ denote the parameters of layer $l$ at step $i$. 
The update equation at step $i$ can be written for each layer as:
\begin{equation}    \label{eq:layerwise_tta}
    \mathbf{\theta}_{i,l} = \mathbf{\theta}_{i-1,l} + \mathbf{u}_{i,l}.
\end{equation}
To perform layer selection, we modify this update equation by introducing a mask: 
\begin{equation}    \label{eq:mask_tta}
    \mathbf{\theta}_{i,l} = \mathbf{\theta}_{i-1,l} + m_{i,l}~\mathbf{u}_{i,l},
\end{equation}
where $m_{i,l}\in\{0,1\}$ is the value of the binary mask applied to the update $\mathbf{u}_{i,l}$.

\subsection{Cosine distance criterion}
\label{subsec:LEASTcosine}
Existing works have shown that gradient descent happens in a tiny subspace \cite{gur2018gradient}. Moreover, as the model reaches closer to the minima, the gradients across the samples get noisy \cite{mahsereci2017early, forouzesh2021disparity, agarwal2022estimating}. We aim to identify the layers with the most beneficial gradient updates to the model for adapting to the new domain. Let us assume that the total displacement of parameters of layer $l$ at the start of the $i^\text{th}$ step is given by: 
\begin{equation}    \label{eq:td}
    \mathbf{TD}_{i-1,l} = \sum_{j=1}^{i-1} m_{j,l}\mathbf{u}_{j,l}=\mathbf{\theta}_{i-1,l} - \mathbf{\theta}_{0,l}.
\end{equation}
Our proposed criterion relies on the angular deviation of the update $\mathbf{u}_{i,l}$ from the direction of the total displacement that would result from making this update:
\begin{equation}    \label{eq:cosine}
\cos(\alpha_{i,l}) =\frac{\mathbf{u}_{i,l} \cdot (\mathbf{u}_{i,l}+\mathbf{TD}_{i-1,l})}{\Vert\mathbf{u}_{i,l}\Vert_2 ~\Vert\mathbf{u}_{i,l}+\mathbf{TD}_{i-1,l}\Vert_2}.
\end{equation}
This angle can be interpreted as the deviation of the update under consideration from the anticipated average update, which has the same direction as the anticipated total displacement $\mathbf{u}_{i,l}+\mathbf{TD}_{i-1,l}$ -- see this illustrated in Fig.\ \ref{fig:methodology}. 

Comparing our criterion across layers allows us to define which update is performed by defining the mask:
\begin{equation}
m_{i,l} = 
\begin{cases} 
1 & \text{if $\cos(\alpha_{i,l}) > \lambda$} \\
0 & \text{otherwise}
\end{cases},
\end{equation}
where $\lambda$ is the selection threshold. The fact that the cosine metric lies in the $[-1,1]$ domain allows us to compare the alignment of updates for layers with different sizes of parameters.
We set a single $\lambda > 0$ for thresholding over all layers, which prevents the adaptation of updates that are misaligned with the updates applied in the past. A $\lambda$ close to $1$ will only allow adaptation of updates aligned with past updates, while a lower $\lambda$ would be less restrictive. 

\subsection{Cosine distance with reset}
\label{subsec:LEASTresetWindow}

While the cosine distance can stop adaptation for noisy gradients, our criterion may fail, especially when the gradient update trajectory needs to change direction after a certain point. If the gradient updates meet an inflection point in the loss landscape, cosine distance will prevent further adaptation, and the model will remain stuck at this point even if the gradient update is informative. To solve such cases, we propose to use resets for the computation of the total displacement of a layer. We use a fixed window scheme for resetting the initial parameter point, which we will call the \emph{anchor point}. This corresponds to: 
\begin{equation}    \label{eq:td_with_reset}
    \mathbf{TD}_{i,l} = \mathbf{\theta}_{i,l} - \mathbf{\theta}_{r,l},
\end{equation}
where $\mathbf{\theta}_{r,l}$ is the parameter at last reset step $r = \lfloor \frac{i-1}{s} \rfloor$, and $s$ is the size of the reset window. The anchor point changes only when the reset window changes, as illustrated in Fig.\ \ref{fig:mask}.

\section{Experiments}
\label{sec:expResults}

This section compares our proposed approaches with existing baselines on Domainbed \cite{gulrajani2021search}, a popular benchmark with large single distribution shifts, and Continual TTA, a popular benchmark with multiple distribution shifts.

\noindent \emph{TTA losses}~~ Two popular TTA losses are considered: Pseudo-Labeling (PL) \cite{lee2013pseudo} and SHOT \cite{liang2020we}. We perform hyperparameter selection based on \citet{zhao2023pitfalls}, where we report the performance for the best hyperparameter set found by sweeping over a range of values.

\noindent\emph{Baselines}~~ We compare the TTA performance obtained by adapting \textit{All layers} vs. the layers proposed by our approach. We also report the \textit{ERM} (\textit{no adaptation}) performance of the pretrained model. In Domainbed, we also compare against AutoRGN and AutoSNR \cite{lee2023surgical}, two popular baselines proposed to identify optimal layers in fine-tuning setup.

\noindent\emph{Implementational details}~~ We report results for \LEAST{} with \emph{window size} of 20 and \emph{selection threshold} of 0.75 with single-layer granularity. It appears that \LEAST{} is not overly sensitive to hyperparameters, and those values work well overall -- see Sec.\ \ref{sec:ablationLEAST} for more discussion on hyperparameter values and the design choices. We also scale the updates for a few initial samples in the reset window to reduce their impact on incorrect layer selection.

\begin{table}[t]
\centering
\resizebox{1.0\linewidth}{!}{%
\begin{tabular}{l@{\hskip 0.5em}l@{\hskip 0.5em}l@{\hskip 0.5em}|@{\hskip 0.5em}c@{\hskip 0.5em}c@{\hskip 0.5em}c@{\hskip 0.5em}c@{\hskip 0.5em}|@{\hskip 0.5em}c}
\toprule
& \textbf{TTA}   & \textbf{Method} & \textbf{PACS} $\uparrow$                  & \textbf{VLCS} $\uparrow$                  & \textbf{Terra} $\uparrow$        & \textbf{Office} $\uparrow$            & \textbf{Mean} $\uparrow$\\
\midrule
\multirow{9}{*}{\rotatebox{90}{ResNet-18}} &
ERM                     & --       & $80.99~(\pm 0.9)$          & $75.14~(\pm 1.2)$          & $40.80~(\pm 0.2)$          & $62.18~(\pm 0.4)$          & $64.78$\\
\cmidrule{2-8}
& \multirow{4}{*}{PL}   & All layers & $81.79~(\pm 0.7)$          & $65.69~(\pm 1.5)$          & $35.40~(\pm 9.7)$          & $60.20~(\pm 1.4)$          & $60.77$\\
&                       & AutoRGN & $82.82~(\pm 0.6)$          & $72.63~(\pm 1.3)$          & $38.18~(\pm 6.1)$          & $62.38~(\pm 0.2)$          & $64.00$\\
&                       & AutoSNR & $80.58~(\pm 1.2)$          & $65.72~(\pm 1.8)$          & $35.01~(\pm 10.4)$         & $59.82~(\pm 0.9)$          & $60.28$\\
&                       & \LEAST{}    & $\mathbf{83.56~(\pm 0.6)}$ & $\mathbf{75.48~(\pm 1.2)}$ & $\mathbf{44.19~(\pm 1.1)}$ & $\mathbf{62.67~(\pm 0.2)}$ & $\mathbf{66.47}$\\
\cmidrule{2-8}
& \multirow{4}{*}{SHOT} & All layers & $83.48~(\pm 0.3)$          & $66.23~(\pm 2.8)$          & $33.81~(\pm 1.3)$          & $63.03~(\pm 0.4)$          & $61.64$\\
&                       & AutoRGN  & $\mathbf{84.10~(\pm 0.5)}$ & $69.78~(\pm 1.3)$          & $37.37~(\pm 0.7)$          & $63.09~(\pm 0.2)$          & $63.59$\\
&                       & AutoSNR  & $83.43~(\pm 0.3)$          & $66.26~(\pm 2.7)$          & $33.75~(\pm 1.2)$          & $63.02~(\pm 0.4)$          & $61.62$\\
&                       & \LEAST{}    & $83.92~(\pm 0.8)$          & $\mathbf{76.23~(\pm 1.1)}$ & $\mathbf{42.13~(\pm 1.4)}$ & $\mathbf{63.32~(\pm 0.3)}$ & $\mathbf{66.40}$\\
\midrule
\multirow{9}{*}{\rotatebox{90}{ResNet-50}}
& ERM                   & --       & $82.84~(\pm 0.5)$          & $75.83~(\pm 0.9)$          & $46.14~(\pm 2.3)$          & $66.93~(\pm 0.3)$          & $67.93$\\
\cmidrule{2-8}
& \multirow{4}{*}{PL}   & All layers & $82.36~(\pm 2.8)$          & $69.22~(\pm 1.4)$          & $42.28~(\pm 3.2)$          & $61.54~(\pm 3.3)$          & $63.85$\\
&                       & AutoRGN  & $\mathbf{85.03~(\pm 1.9)}$ & $75.35~(\pm 1.4)$          & $48.44~(\pm 2.4)$          & $66.93~(\pm 0.3)$          & $68.94$\\
&                       & AutoSNR  & $83.41~(\pm 3.4)$          & $70.14~(\pm 4.6)$          & $44.08~(\pm 3.4)$          & $61.95~(\pm 3.0)$          & $64.90$\\
&                       & \LEAST{}    & $84.87~(\pm 0.8)$          & $\mathbf{76.88~(\pm 1.6)}$ & $\mathbf{50.10~(\pm 2.5)}$  & $\mathbf{67.34~(\pm 0.3)}$ & $\mathbf{69.80}$\\
\cmidrule{2-8}
& \multirow{4}{*}{SHOT} & All layers & $85.15~(\pm 1.1)$          & $64.25~(\pm 1.1)$          & $35.33~(\pm 3.1)$          & $67.37~(\pm 0.3)$          & $63.03$\\
&                       & AutoRGN & $\mathbf{86.34~(\pm 1.1)}$ & $70.2~(\pm 0.9)$           & $40.59~(\pm 1.3)$          & $68.10~(\pm 0.4)$          & $66.31$\\
&                       & AutoSNR & $85.51~(\pm 0.5)$          & $64.26~(\pm 1.3)$          & $34.97~(\pm 3.2)$          & $67.33~(\pm 0.2)$          & $63.02$\\
&                       & \LEAST{}    & $86.13~(\pm 0.8)$          & $\mathbf{76.48~(\pm 1.0)}$ & $\mathbf{45.94~(\pm 1.6)}$ & $\mathbf{68.13~(\pm 0.3)}$ & $\mathbf{69.17}$\\
\bottomrule
\end{tabular}%
}
\caption{Accuracy (\%) of various layer selection methods on Domainbed benchmark (setup described in Sec.\ \ref{subsec:domainbed}). The best method for a given TTA loss and backbone is in bold.}
\label{tab:dg_bench_results}
\end{table}

\subsection{Domainbed results}
\label{subsec:domainbed}

For the experiments on Domainbed, we follow the evaluation protocol as described in \citet{iwasawa2021test}, including dataset splits for the following four datasets: PACS \cite{li2017deeper}, VLCS \cite{fang2013unbiased}, Terra Incognita \cite{beery2018recognition}, and Office-Home \cite{venkateswara2017deep}. Results are reported on two backbones (i.e., ResNet-18 and ResNet-50) with batch normalization layers, while the pretrained models are made using default hyperparameters described in \citet{gulrajani2021search}. Mean and standard deviation are reported over three repetitions with different random seeds. See Appendix \ref{supp:sec:domainbed} for further details.

Key takeaways from results are reported in Tab.\ \ref{tab:dg_bench_results}:
\begin{itemize}
\item \LEAST{} outperforms ERM (no adaptation) by 2\% overall and \textit{All layers} TTA baselines by more than 5\% overall across all losses, backbones, and datasets.
\item Existing layer selection baselines like AutoRGN or AutoSNR can improve performance compared to all layers TTA in most setups, especially AutoRGN, but fail to improve against no adaptation baselines for some datasets like VLCS or TerraIncognita or some TTA losses like SHOT. \LEAST{} consistently demonstrates equivalent or superior performance across all datasets and TTA losses, achieving an overall improvement of about 2\%.
\item \LEAST{} improves over Domainbed large shift datasets (i.e., PACS, OfficeHome) similar to AutoRGN and AutoSNR while comfortably outperforming the ERM baseline. On small shift datasets (i.e., VLCS, TerraIncognita), existing baselines struggle to outperform the \textit{no adaptation} baseline while \LEAST{} appears to prevent degradation caused by over-adaptation, thereby enhancing performance over the ERM baseline and safeguard against further degradation.
\end{itemize}

\begin{table}[t]
\centering
\resizebox{0.65\columnwidth}{!}{%
\begin{tabular}{ll|cc}
\toprule
\textbf{TTA}                   & \textbf{Method}               & \textbf{CIFAR10C} $\downarrow$        & \textbf{CIFAR100C} $\downarrow$ \\
\midrule
ERM                &  & $43.50~(\pm 18.7)$          & $46.40~(\pm 15.7)$   \\
\midrule
\multirow{3}{*}{PL}   & All layers           & $88.72~(\pm 1.2)$          & $98.63~(\pm 1.5)$   \\
                      & \LEAST{}               & $\mathbf{28.68~(\pm 6.6)}$ & $\mathbf{33.69~(\pm 5.7)}$ \\
\midrule
\multirow{3}{*}{SHOT} & All layers           & $89.33~(\pm 2.3)$          & $97.32~(\pm4.8)$          \\
                      & \LEAST{}                & $\mathbf{20.46~(\pm 7.7)}$ & $\mathbf{32.87~(\pm 5.6)}$ \\
\bottomrule
\end{tabular}%
}
\caption{Accuracy (\%) of layer selection methods on Continual TTA benchmark (with the setup described in Sec. \ref{subsec:cotta}). The best method for a given TTA loss is in bold.}
\label{tab:cotta}
\end{table}

\subsection{Continual TTA results}
\label{subsec:cotta}

We follow the evaluation protocol as described in \citet{wang2022continual},  evaluating performance on two datasets-backbones: 1) CIFAR10C \cite{hendrycks2019benchmarking} with WideResNet-28 \cite{zagoruyko2016wide} and CIFAR100C \cite{hendrycks2019benchmarking} with ResNeXt-29 \cite{xie2017aggregated}. The pretrained models are trained as described in Robustbench \cite{croce2robustbench}. Mean and standard deviation are reported across the 15 corruption types. Further details are given in Appendix \ref{supp:sec:continualTTA}.

The key takeaways based on the results from Tab. \ref{tab:cotta} are:
\begin{itemize}
\item Performance degradation by training all layers is worse in the Continual TTA benchmark containing multi-domain shifts than degradation in the Domainbed benchmark containing single-domain shifts. Moreover, more severe degradations are observed in CIFAR100C, which has 100 classes, compared to CIFAR10, which includes 10 classes, despite similar ERM performance on both datasets.
\item \LEAST{} consistently outperforms ERM by about 15\% and all layers TTA baseline by about 65\%,  despite severe degradation.
\end{itemize}

\begin{figure*}[t]
  \centering
  \begin{subfigure}{0.45\linewidth}
    \centering
    \includegraphics[scale=0.325]{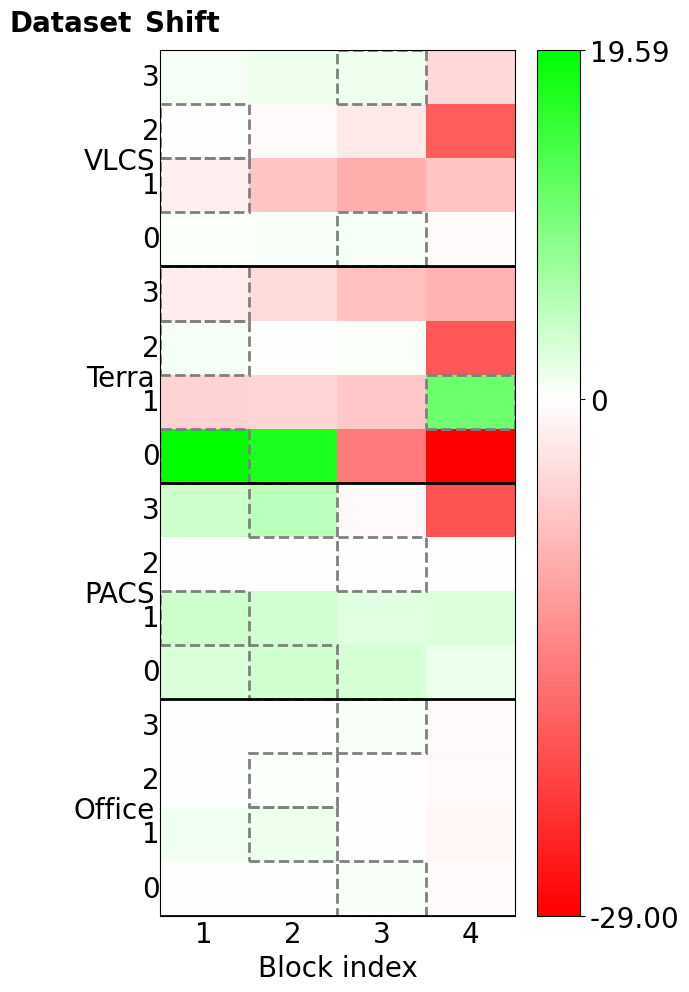}
    \caption{PL loss-based TTA}
    \label{fig:res18_PL}
  \end{subfigure}
  \begin{subfigure}{0.45\linewidth}
    \centering
    \includegraphics[scale=0.325]{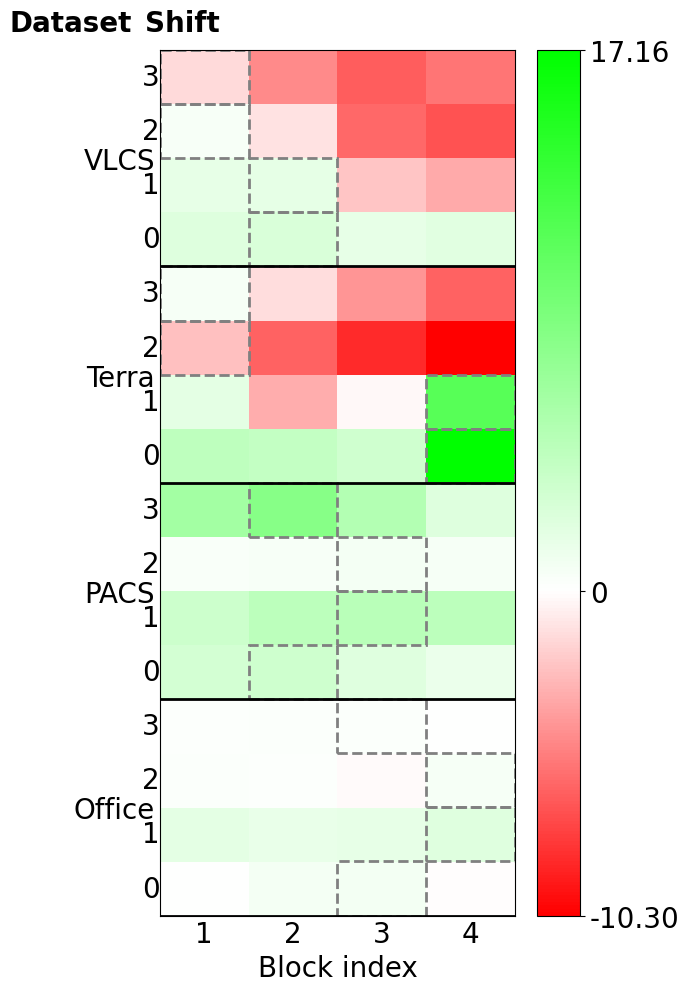}
    \caption{SHOT loss-based TTA}
    \label{fig:res18_SHOT}
  \end{subfigure}
  \caption{Heatmap of Performance improvement (\%) per-block on Domainbed benchmark. Performance improvement is the difference between the TTA accuracy of a given block/layer and ERM accuracy for the same shift. Positive performance improvements are shown in green, and negative performance improvements (or degradation) are in red. Using the bounding box, we highlight the best block per loss and dataset shift. Further details in Sec. \ref{sec:study}.}
  \label{fig:heatmap}
\end{figure*}

\section{Layer Selection Study}
\label{sec:study}
In this section, we evaluate the importance of layer selection for test time adaptation on the Domainbed benchmark and provide some analysis and motivation for \LEAST{}. We use the Domainbed benchmark with the ResNet-18 backbone, which contains four blocks of layers. We study the effect of choosing one block over another by performing adaptation on a single block while freezing all the other blocks of the model. We refer to blocks and layers interchangeably in this section. We report the difference between TTA and ERM accuracy over all blocks for each loss and dataset shift setting. Otherwise, we rely on the same setup and evaluation protocol described in Sec.\ \ref{subsec:domainbed}.

\subsection{Layer selection matters}

In Fig.\ \ref{fig:heatmap}, we observe that not all layers are equally receptive to adaptation. We refer to a layer as \good{} or \bad{} based on the accuracy improvement of selecting a given layer w.r.t. performance of a pretrained model on the same shift.  We compare against Empirical Risk Minimization (ERM) or the frozen pretrained model's performances, as we are interested in measuring the performance improvement or degradation brought by individual layers during adaptation. Also, the ERM model performs better on average than \textit{all layer} TTA, as seen in Sec.\ \ref{sec:expResults}, and it becomes a natural baseline that can help contrast different layers.

The selection of layers in existing TTA approaches typically remains unchanged in all adaptation settings. We find it can be a suboptimal strategy and one of the major causes of degradation in existing TTA approaches -- no single layer adaptation is suitable for all settings. Therefore, layer selection is essential for TTA, and we propose \LEAST{} to improve the performance of existing TTA approaches in various settings.

\subsection{What affects the adaptability of a layer?}
Using the same setup and evaluation protocol (cf., Sec.\ \ref{subsec:domainbed}), we are making the following observations on Fig. \ref{fig:heatmap} about the factors affecting the adaptability of \good{} layers:
\begin{itemize}
\item Location of \good{} layers in a model can change across shifts of a given dataset, despite using pretrained models trained on the same class labels. Similar observations have also been made in fine-tuning setups\cite{lee2023surgical}. There is a need for a good layer selection criterion that depends on target samples observed by the model at test time.
\item We also find that \good{} layers in a model can change with different TTA loss functions, even for the same shift and dataset. Hence, a good layer selection criterion must also depend on the TTA loss function used to adapt the model at inference.
\end{itemize}
Since gradients depend on the shift and TTA loss function used, \LEAST{} uses layerwise gradients to identify the adaptability of each layer in the model.

\subsection{How do \good{} layers differ from \bad{} layers?}
\label{subsec:goodvsbad}

To perform a detailed per-layer analysis, we created the Tiny-Domainbed benchmark, which was made as a smaller version of Domainbed. It consists of all the critical shifts with the brightest red/green layers (displayed in Fig.\ \ref{fig:heatmap}), whose \good{} layers can also change with the TTA method. We follow the benchmark and evaluation protocol described in Sec.\ \ref{subsec:domainbed}, with further details given in Appendix \ref{supp:sec:tinydomainbed}. Based on Tab.\ \ref{tab:study}, the following are the differences between \good{} and \bad{} layers:
\begin{itemize}
\item Adaptation with \textit{Worst Block} results in poor TTA accuracy, poorer generalization to the target domain, and higher forgetting. Since training all layers involves training the worst layer, this could explain why training all layers results in poorer TTA accuracy. On the other hand, \textit{Best Block} results in better generalization to the target domain. This implies that TTA with \good{} layers can potentially learn target domain features better than TTA with \bad{} layers.

\item We observe that \textit{Best Block} results in reduced source forgetting compared to \textit{Worst Block}. This implies that TTA with \good{} layers strikes an improved balance between learning new features on the target domain while retaining useful pretrained features from the source domain.
\end{itemize}
Therefore, we propose \LEAST{} to identify \good{} layers for adaptation, which can help balance adaptation to the new domain while reducing source forgetting.

\begin{table}[t]
\centering

\resizebox{\linewidth}{!}{%
\begin{tabular}{l|ccccc}
\toprule
\textbf{Method}   & \textbf{TTA Acc.} $\uparrow$ & \textbf{Gen.} $\uparrow$ & \textbf{Forget.}  $\downarrow$ & \textbf{Rank corr.} $\uparrow$ \\
\midrule
All Blocks           & 53.6 & 46.5  & 31.3   & N/A  \\
\midrule
Worst Block (oracle) & 43.5 & 38.7  & 39.9  & -1  \\
Best Block (oracle)  & \textbf{64.1} & \textbf{63.9}  & 28.7  & \textbf{1}  \\
\midrule

Random Block         & 53.1 & 49.1  & \underline{13.1}   & 0 \\
\LEAST{}               & \underline{59.4} & \underline{58.0}  &  \textbf{9.3} &  \underline{0.76}   \\
\bottomrule
\end{tabular}%
}
\caption{Effect of various layer selection methods on TTA Accuracy (\%), Generalization (\%), Forgetting (\%) and Spearman correlation with Best Block ($\in[-1,1]$) averaged over different shifts on Tiny-Domainbed benchmark (with the setup described in Sec.\ \ref{subsec:goodvsbad}). \textit{TTA Acc} is the accuracy of testing samples from the target domain seen during adaptation. \textit{Generalization} is the accuracy of the held-out split of the target domain after adaptation. \textit{Forgetting} is the drop in accuracy on the held-out split of source domains after adaptation. \textit{Rank correlation} is the Spearman correlation of layer selection rank between the oracle and the method. Bold and underlined denote best and second-best, respectively.}
\label{tab:study}
\end{table}

\subsection{How does \LEAST{} compare to oracle strategies?}

To analyze \LEAST{}'s layer selection behavior, we compare it to the oracle strategies given by Best block and Worst block on the Tiny-Domainbed benchmark (Tab.\ \ref{tab:study}).

\noindent \textbf{\LEAST{} well approximates the oracle layer selection} \quad
\LEAST{} substantially improves over \textit{All Blocks}, \textit{Worst Block}, and \textit{Random Block} method.  In some sense, \textit{Best Block} method acts as an empirical upper-bound performance if we have access to a target domain with labels while incurring the high computational cost of brute forcing over individual layers of the model. \LEAST{} comes close to this upper bound performance without requiring any target labels using a cheap layer selection criterion. As a result, \LEAST{} also effectively balances computational cost with performance.

\begin{table}[t]
\small
\centering
\resizebox{0.65\linewidth}{!}{%
\begin{tabular}{lc|c}
\toprule
\textbf{Setting}                         & \textbf{Condition} & \textbf{Accuracy} $\uparrow$ \\
\midrule
\multirow{3}{*}{\textbf{Partitioning}}   & Single block       & $67.64$             \\
                                         & Single layer       & $68.57$             \\
                                         & Multiple layers    & $66.48$             \\
\midrule
\multirow{3}{*}{\textbf{Window Size}}    & 5                  & $68.46$             \\
                                         & 20                 & $68.57$             \\
                                         & $\infty$           & $68.37$             \\
\midrule
\multirow{3}{*}{\textbf{Threshold}}      & 0.5                & $68.57$             \\
                                         & 0.75               & $68.57$             \\
                                         & 0.99               & $68.6$              \\
\midrule
\multirow{2}{*}{\textbf{Batch Size = 1}} & All Layers         & $33.47$             \\
                                         & GALA               & $67.28$             \\
\midrule
\midrule
\multirow{2}{*}{\textbf{Continual TTA}}  & No Reset           & $69.9$              \\
                                         & With Reset         & $71.1$             \\
\bottomrule
\end{tabular}%
}
\caption{Accuracy (\%) under different experimental conditions. The values are averaged on Domainbed for the first four settings and Continual TTA for the last.}
\label{tab:analysis}
\end{table}

\begin{figure*}[t]
\centering
\includegraphics[width=0.45\linewidth]{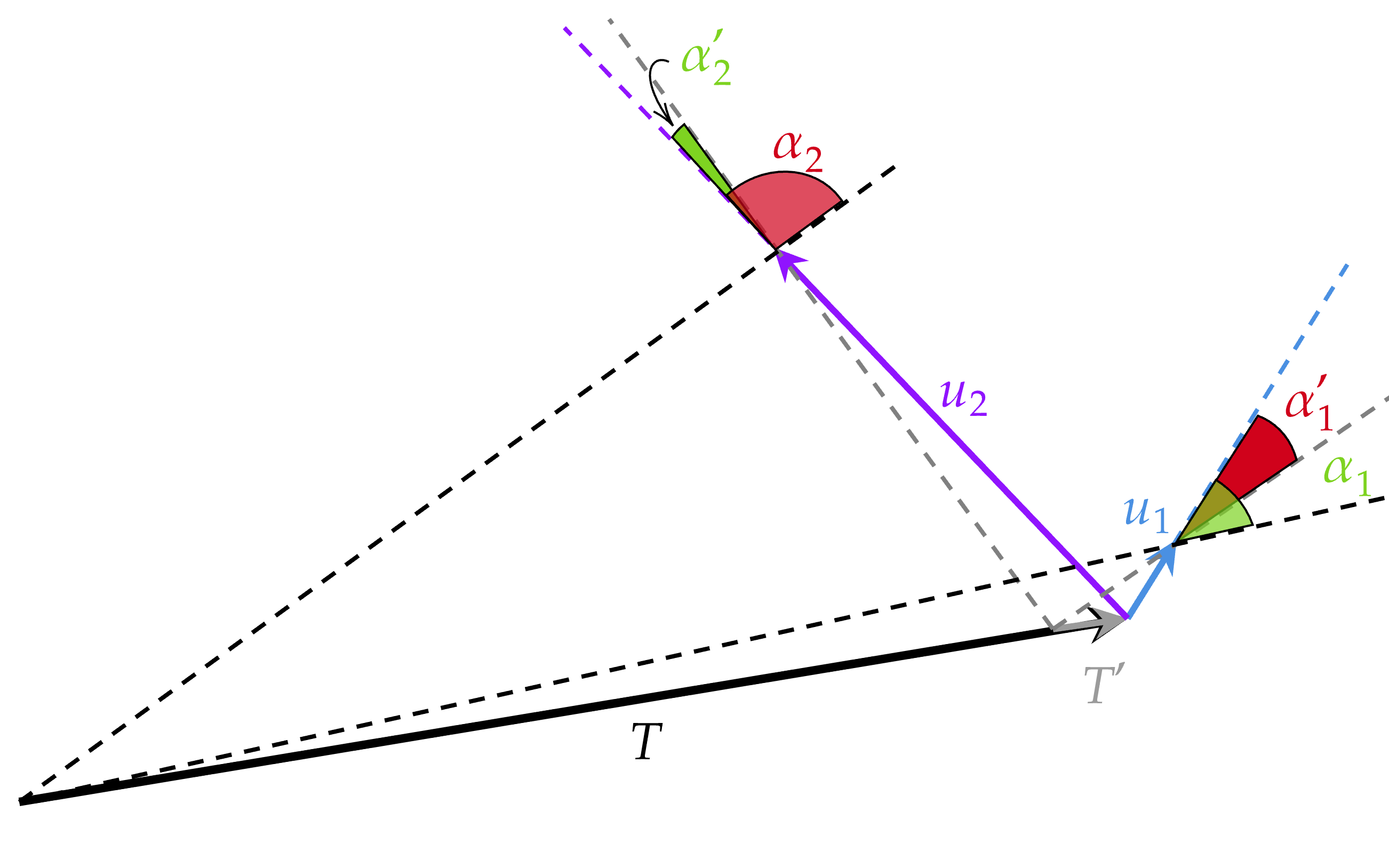}
\includegraphics[width=0.24\linewidth]{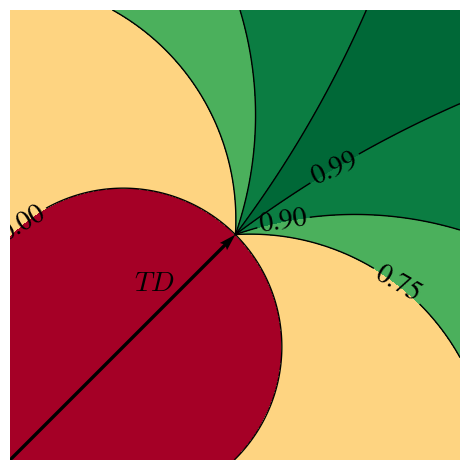}
\includegraphics[width=0.24\linewidth]{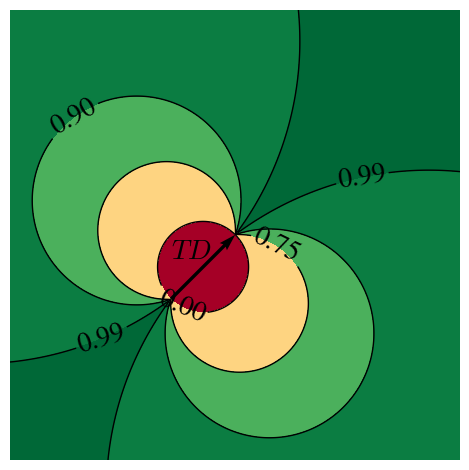}
  \caption{Effect of magnitude of $u$ on cosine distance criterion. \textbf{Left:} Consider two vectors such that $u_1$ is smaller than $u_2$ but is better aligned with its displacement. For large displacements ($T$), alignment becomes crucial and \LEAST{} selects $u_2$. For small displacements ($T^{'}$), the update’s magnitude can dominate the criterion, and \LEAST{} selects $u_1$.
\textbf{Middle and Right:} Plot of cosine metric values with level curves. Alignment prevails for small updates compared to the total displacement (Middle). But, for updates with large magnitude compared to total displacement (Right), large cosine values can be obtained even for misaligned updates.}
  \label{fig:align_vs_magn}
\end{figure*}

\noindent \textbf{\LEAST{} is more conservative than the oracle} \quad
\LEAST{} selects the layers for adaptation with the most aligned gradients. It can stop adaptation if the gradients are noisy or no longer aligned to prevent further degradation. In Tab.\ \ref{tab:study}, we see that it may have aggressively stopped a few useful updates compared to the \textit{Best Block}, our empirical upper bound. As a result, it gets much better at avoiding forgetting but is a bit lower on TTA accuracy and generalization.

\noindent \textbf{\LEAST{} tends to select more often the blocks with better accuracy} \quad
Oracle TTA performance, as measured in Fig.\ \ref{fig:heatmap}, ranks the four blocks for each configuration. Similarly, \LEAST{} chooses to update each layer with a particular frequency during TTA, leading to a ranking of the four blocks. We assess the relationship between these two different ways of ranking blocks using Spearman rank correlation and find $\rho=0.76$ (cf. Tab. \ref{tab:study}), which seems to indicate that the selection strategy used by \LEAST{} is a good proxy for the oracle TTA performance achieved when adapting always the same layer.

\section{Analysis of \LEAST{}}
\label{sec:analysisLEAST}
\label{sec:ablationLEAST}

In this section, we evaluate the impact of different design choices and hyperparameters of \LEAST{} in Tab.\ \ref{tab:analysis}, supporting choices presented in Tab.\ \ref{tab:dg_bench_results}. For the partitioning setting, \emph{Single block} means a single block of many layers is updated at each iteration, \emph{Single layer} corresponds to the best layer selected for the update, and \emph{Multiple layers} corresponds to individually best layers selected for the update based on the cosine distance and the threshold. Also, a window size of $\infty$ implies no reset. Some important observations stemming from Tab.\ \ref{tab:analysis}:
\begin{itemize}
\item Layer granularity performs better than block granularity. At layer granularity, \LEAST{} has better fine-grained control over choosing the layers to adapt, improving performances in all cases tested.

\item Adaptation with the best single layer is much better than with the best multiple layers. Cosine distance can correctly identify the single best layer to train, although it may still struggle to determine the best set of multiple layers to update.

\item Optimal reset-window size can improve performance. We see that a reset window size of 20 works reasonably well across the backbones and the TTA losses tested on Domainbed.

\item The choice of selection threshold is not very sensitive. A threshold of 0.75 seems to work across the board without being too restrictive.
\end{itemize}

In the following section, we briefly analyze some aspects of the proposed approach.

\paragraph{Proposed cosine distance criterion effectively balances gradient magnitude and direction.} 
Let us first rewrite the \LEAST{} criterion in Eq.\ \ref{eq:cosine} for a given layer $l$ in terms of $T=\Vert \textbf{TD}_{i-1, l}\Vert$, $u=\Vert\textbf{u}_{i,l} \Vert$ and the angle $\beta$ between $\textbf{TD}_{i-1,l}$ and $\textbf{u}_{i,l}$. Using the Pythagorean theorem, we obtain:
\begin{align}   \label{eq:cosAlphaBeta_new}
\cos(\alpha) & = \frac{T \cos(\beta) + u}{\sqrt{(T+u \cos(\beta))^2+(u\sin(\beta))^2}}.
\end{align}
We observe that our criterion depends on the norm $T$ of the total displacement, the norm $u$ of the update, and their alignment, given by the angle $\beta$ between these vectors. Fig.\ \ref{fig:align_vs_magn} shows the cosine metric plots. We see that while alignment is crucial for large displacements, 
the update’s magnitude can also dominate for small displacements. For example, consider two layers with the same norm $T$ but different updates $\textbf{u}_1$ and $\textbf{u}_2$.  If $\Vert \textbf{u}_1\Vert$ is smaller than $\Vert \textbf{u}_2\Vert$ but $\textbf{u}_1$ is more aligned with its displacement, two scenarios arise:
\begin{enumerate}
    \item For larger $T$,  \LEAST{} selects layer 1, favoring the alignment and exploiting the learned direction. This scenario would seem more common during TTA.

    \item For small $T$, \LEAST{} selects layer 2, favoring the magnitude, and can explore over different directions. This can occur for initial samples.
\end{enumerate}
Consequently, \LEAST{} effectively balances the gradient magnitude and the direction of gradients for selecting the best layer. More discussion is in Appendix \ref{supp:sec:LEASTdiscussion}.

\paragraph{Proposed layer selection framework offers a more flexible adaptation strategy for TTA.}~~ The selection of layers in existing TTA approaches typically remains unchanged across different shifts. On the other hand, sample selection-based TTA \cite{niu2022efficient} approaches aim to improve performance by skipping the adaptation of all layers on a few unreliable samples. Based on Eq.\ \ref{eq:mask_tta} and Fig.\ \ref{supp:fig:ls_vs_es_les}
,  we can see that \LEAST{} is more flexible and general than the existing layer selection and sample selection strategies in TTA for performing layerwise adaptation.

\paragraph{Reset mechanism seems beneficial in multi-domain shift settings.}~~ Comparing \LEAST{} with and without reset on Tab.\ \ref{tab:analysis}, we see that while reset yields only marginal improvement on Domainbed, a single-domain shift benchmark, its benefits are more evident on a multi-shift benchmark like Continual TTA. This indicates that the reset mechanism's ability to facilitate slight adjustments in the overall gradient update direction may be advantageous in a continuously changing testing domain.

\paragraph{\LEAST{} is quite robust on single sample adaptation.}~~ In Tab. \ref{tab:analysis}, we show that in the adverse setting of batch size of 1, while existing TTA approaches witness severe performance degradation, \LEAST{} improves on \textit{all layers} baseline on Domainbed.

\section{Conclusion}
\label{sec:conc}

In this paper, we introduce Gradient Aligned Layer Adaptation (GALA), a novel layer selection framework explicitly designed for Test Time Adaptation (TTA). Our comprehensive study reveals that layers in neural networks exhibit varying receptiveness to adaptation, and the optimal set of layers for adaptation depends on both the specific distribution shift and the loss function employed during inference. Building on these insights, we propose GALA, a dynamic layer selection criterion that ranks layers based on gradient alignment, effectively mitigating overfitting and performance degradation. Extensive experiments across diverse datasets, model architectures, and TTA losses demonstrate GALA's superior performance compared to existing methods, including standard \textit{ERM}, \textit{all-layers} adaptation, and other layer selection baselines.


\section*{Acknowledgments}
This work is supported by the DEEL Project CRDPJ 537462-18 funded by the Natural Sciences and Engineering Research Council of Canada (NSERC) and the Consortium for Research and Innovation in Aerospace in Québec (CRIAQ), together with its industrial partners Thales Canada inc, Bell Textron Canada Limited, CAE inc and Bombardier inc. \footnote{\url{https://deel.quebec}}
Computations were made on the cedar, and beluga supercomputers, managed by Calcul Québec and the Digital Research Alliance of Canada (Alliance). We extend our gratitude to the members of the \#lunch-at-mila and \#deel\_ood for their valuable input, with special thanks to Vineetha Kondameedi for her essential feedback in enhancing the quality of this paper.

\bibliography{references}

\begin{thebibliography}{93}
\providecommand{\natexlab}[1]{#1}

\bibitem[{Agarwal, D'souza, and Hooker(2022)}]{agarwal2022estimating}
Agarwal, C.; D'souza, D.; and Hooker, S. 2022.
\newblock Estimating example difficulty using variance of gradients.
\newblock In \emph{Proceedings of the IEEE/CVF Conference on Computer Vision and Pattern Recognition}, 10368--10378.

\bibitem[{Ahn, Kim, and Oh(2019)}]{ahn2019deep}
Ahn, C.; Kim, E.; and Oh, S. 2019.
\newblock Deep elastic networks with model selection for multi-task learning.
\newblock In \emph{Proceedings of the IEEE/CVF international conference on computer vision}, 6529--6538.

\bibitem[{Andriushchenko and Flammarion(2020)}]{andriushchenko2020understanding}
Andriushchenko, M.; and Flammarion, N. 2020.
\newblock Understanding and improving fast adversarial training.
\newblock \emph{Advances in Neural Information Processing Systems}, 33: 16048--16059.

\bibitem[{Bai et~al.(2021)Bai, Yang, Han, Yang, Li, Mao, Niu, and Liu}]{bai2021understanding}
Bai, Y.; Yang, E.; Han, B.; Yang, Y.; Li, J.; Mao, Y.; Niu, G.; and Liu, T. 2021.
\newblock Understanding and improving early stopping for learning with noisy labels.
\newblock \emph{Advances in Neural Information Processing Systems}, 34: 24392--24403.

\bibitem[{Barba, Jaggi, and Dandi(2021)}]{barba2021implicit}
Barba, L.; Jaggi, M.; and Dandi, Y. 2021.
\newblock Implicit gradient alignment in distributed and federated learning.
\newblock In \emph{AAAI Conference on Artificial Intelligence, AAAI}, volume~22.

\bibitem[{Beery, Van~Horn, and Perona(2018)}]{beery2018recognition}
Beery, S.; Van~Horn, G.; and Perona, P. 2018.
\newblock Recognition in terra incognita.
\newblock In \emph{Proceedings of the European conference on computer vision (ECCV)}, 456--473.

\bibitem[{Bonet et~al.(2021)Bonet, Ortega, Ruiz-Hidalgo, and Shekkizhar}]{bonet2021channel}
Bonet, D.; Ortega, A.; Ruiz-Hidalgo, J.; and Shekkizhar, S. 2021.
\newblock Channel-wise early stopping without a validation set via NNK polytope interpolation.
\newblock In \emph{2021 Asia-Pacific Signal and Information Processing Association Annual Summit and Conference (APSIPA ASC)}, 351--358. IEEE.

\bibitem[{Bordes et~al.(2023)Bordes, Balestriero, Garrido, Bardes, and Vincent}]{bordes2023guillotine}
Bordes, F.; Balestriero, R.; Garrido, Q.; Bardes, A.; and Vincent, P. 2023.
\newblock Guillotine Regularization: Why removing layers is needed to improve generalization in Self-Supervised Learning.
\newblock \emph{Transactions on Machine Learning Research}.

\bibitem[{Boudiaf et~al.(2022)Boudiaf, Mueller, Ben~Ayed, and Bertinetto}]{boudiaf2022parameter}
Boudiaf, M.; Mueller, R.; Ben~Ayed, I.; and Bertinetto, L. 2022.
\newblock Parameter-free online test-time adaptation.
\newblock In \emph{Proceedings of the IEEE/CVF Conference on Computer Vision and Pattern Recognition}, 8344--8353.

\bibitem[{Burns and Steinhardt(2021)}]{burns2021limitations}
Burns, C.; and Steinhardt, J. 2021.
\newblock Limitations of post-hoc feature alignment for robustness.
\newblock In \emph{Proceedings of the IEEE/CVF Conference on Computer Vision and Pattern Recognition}, 2525--2533.

\bibitem[{Chattopadhyay, Balaji, and Hoffman(2020)}]{2020EccvDMG}
Chattopadhyay, P.; Balaji, Y.; and Hoffman, J. 2020.
\newblock Learning to Balance Specificity and Invariance for In and Out of Domain Generalization.
\newblock In \emph{European Conference in Computer Vision (ECCV)}.

\bibitem[{Chen et~al.(2022)Chen, Wang, Darrell, and Ebrahimi}]{chen2022contrastive}
Chen, D.; Wang, D.; Darrell, T.; and Ebrahimi, S. 2022.
\newblock Contrastive test-time adaptation.
\newblock In \emph{Proceedings of the IEEE/CVF Conference on Computer Vision and Pattern Recognition}, 295--305.

\bibitem[{Choi et~al.(2010)Choi, Lim, Torralba, and Willsky}]{choi2010exploiting}
Choi, M.~J.; Lim, J.~J.; Torralba, A.; and Willsky, A.~S. 2010.
\newblock Exploiting hierarchical context on a large database of object categories.
\newblock In \emph{2010 IEEE computer society conference on computer vision and pattern recognition}, 129--136. IEEE.

\bibitem[{Croce et~al.(2021)Croce, Andriushchenko, Sehwag, Debenedetti, Flammarion, Chiang, Mittal, and Hein}]{croce2robustbench}
Croce, F.; Andriushchenko, M.; Sehwag, V.; Debenedetti, E.; Flammarion, N.; Chiang, M.; Mittal, P.; and Hein, M. 2021.
\newblock RobustBench: a standardized adversarial robustness benchmark.
\newblock In \emph{Thirty-fifth Conference on Neural Information Processing Systems Datasets and Benchmarks Track (Round 2)}.

\bibitem[{Darrin et~al.(2024)Darrin, Staerman, Gomes, Cheung, Piantanida, and Colombo}]{darrin2024unsupervised}
Darrin, M.; Staerman, G.; Gomes, E. D.~C.; Cheung, J.~C.; Piantanida, P.; and Colombo, P. 2024.
\newblock Unsupervised layer-wise score aggregation for textual ood detection.
\newblock In \emph{Proceedings of the AAAI Conference on Artificial Intelligence}, volume~38, 17880--17888.

\bibitem[{Du et~al.(2018)Du, Czarnecki, Jayakumar, Farajtabar, Pascanu, and Lakshminarayanan}]{du2018adapting}
Du, Y.; Czarnecki, W.~M.; Jayakumar, S.~M.; Farajtabar, M.; Pascanu, R.; and Lakshminarayanan, B. 2018.
\newblock Adapting auxiliary losses using gradient similarity.
\newblock \emph{arXiv preprint arXiv:1812.02224}.

\bibitem[{ElAraby et~al.(2023)ElAraby, Sahoo, Pequignot, Novello, and Paull}]{elaraby2023grood}
ElAraby, M.; Sahoo, S.; Pequignot, Y.; Novello, P.; and Paull, L. 2023.
\newblock GROOD: GRadient-aware Out-Of-Distribution detection in interpolated manifolds.
\newblock \emph{arXiv preprint arXiv:2312.14427}.

\bibitem[{Everingham et~al.(2010)Everingham, Van~Gool, Williams, Winn, and Zisserman}]{everingham2010pascal}
Everingham, M.; Van~Gool, L.; Williams, C.~K.; Winn, J.; and Zisserman, A. 2010.
\newblock The pascal visual object classes (voc) challenge.
\newblock \emph{International journal of computer vision}, 88: 303--338.

\bibitem[{Fang, Xu, and Rockmore(2013)}]{fang2013unbiased}
Fang, C.; Xu, Y.; and Rockmore, D.~N. 2013.
\newblock Unbiased metric learning: On the utilization of multiple datasets and web images for softening bias.
\newblock In \emph{Proceedings of the IEEE International Conference on Computer Vision}, 1657--1664.

\bibitem[{Fei-Fei, Fergus, and Perona(2004)}]{fei2004learning}
Fei-Fei, L.; Fergus, R.; and Perona, P. 2004.
\newblock Learning generative visual models from few training examples: An incremental bayesian approach tested on 101 object categories.
\newblock In \emph{2004 conference on computer vision and pattern recognition workshop}, 178--178. IEEE.

\bibitem[{Forouzesh and Thiran(2021)}]{forouzesh2021disparity}
Forouzesh, M.; and Thiran, P. 2021.
\newblock Disparity between batches as a signal for early stopping.
\newblock In \emph{Machine Learning and Knowledge Discovery in Databases. Research Track: European Conference, ECML PKDD 2021, Bilbao, Spain, September 13--17, 2021, Proceedings, Part II 21}, 217--232. Springer.

\bibitem[{Fort et~al.(2019)Fort, Nowak, Jastrzebski, and Narayanan}]{fort2019stiffness}
Fort, S.; Nowak, P.~K.; Jastrzebski, S.; and Narayanan, S. 2019.
\newblock Stiffness: A new perspective on generalization in neural networks.
\newblock \emph{arXiv preprint arXiv:1901.09491}.

\bibitem[{Gao et~al.(2023)Gao, Zhang, Liu, Darrell, Shelhamer, and Wang}]{gao2023back}
Gao, J.; Zhang, J.; Liu, X.; Darrell, T.; Shelhamer, E.; and Wang, D. 2023.
\newblock Back to the source: Diffusion-driven adaptation to test-time corruption.
\newblock In \emph{Proceedings of the IEEE/CVF Conference on Computer Vision and Pattern Recognition}, 11786--11796.

\bibitem[{Gao et~al.(2021)Gao, Zhang, Huang, Wang, and Zhong}]{gao2021gradient}
Gao, Z.; Zhang, S.; Huang, K.; Wang, Q.; and Zhong, C. 2021.
\newblock Gradient distribution alignment certificates better adversarial domain adaptation.
\newblock In \emph{Proceedings of the IEEE/CVF international conference on computer vision}, 8937--8946.

\bibitem[{Gauch et~al.(2022)Gauch, Beck, Adler, Kotsur, Fiel, Eghbal-zadeh, Brandstetter, Kofler, Holzleitner, Zellinger et~al.}]{gauch2022few}
Gauch, M.; Beck, M.; Adler, T.; Kotsur, D.; Fiel, S.; Eghbal-zadeh, H.; Brandstetter, J.; Kofler, J.; Holzleitner, M.; Zellinger, W.; et~al. 2022.
\newblock Few-shot learning by dimensionality reduction in gradient space.
\newblock In \emph{Conference on Lifelong Learning Agents}, 1043--1064. PMLR.

\bibitem[{Gidaris, Singh, and Komodakis(2018)}]{gidaris2018unsupervised}
Gidaris, S.; Singh, P.; and Komodakis, N. 2018.
\newblock Unsupervised representation learning by predicting image rotations.
\newblock \emph{arXiv preprint arXiv:1803.07728}.

\bibitem[{Gong et~al.(2022)Gong, Jeong, Kim, Kim, Shin, and Lee}]{gong2022note}
Gong, T.; Jeong, J.; Kim, T.; Kim, Y.; Shin, J.; and Lee, S.-J. 2022.
\newblock Note: Robust continual test-time adaptation against temporal correlation.
\newblock \emph{Advances in Neural Information Processing Systems}, 35: 27253--27266.

\bibitem[{Gulrajani and Lopez-Paz(2021)}]{gulrajani2021search}
Gulrajani, I.; and Lopez-Paz, D. 2021.
\newblock In Search of Lost Domain Generalization.
\newblock In \emph{International Conference on Learning Representations}.

\bibitem[{Guo, Lee, and Ulbricht(2020)}]{guo2020learning}
Guo, P.; Lee, C.-Y.; and Ulbricht, D. 2020.
\newblock Learning to branch for multi-task learning.
\newblock In \emph{International conference on machine learning}, 3854--3863. PMLR.

\bibitem[{Guo et~al.(2019)Guo, Shi, Kumar, Grauman, Rosing, and Feris}]{guo2019spottune}
Guo, Y.; Shi, H.; Kumar, A.; Grauman, K.; Rosing, T.; and Feris, R. 2019.
\newblock Spottune: transfer learning through adaptive fine-tuning.
\newblock In \emph{Proceedings of the IEEE/CVF conference on computer vision and pattern recognition}, 4805--4814.

\bibitem[{Gupta, Yadav, and Paull(2020)}]{gupta2020look}
Gupta, G.; Yadav, K.; and Paull, L. 2020.
\newblock Look-ahead meta learning for continual learning.
\newblock \emph{Advances in Neural Information Processing Systems}, 33: 11588--11598.

\bibitem[{Gur-Ari, Roberts, and Dyer(2018)}]{gur2018gradient}
Gur-Ari, G.; Roberts, D.~A.; and Dyer, E. 2018.
\newblock Gradient descent happens in a tiny subspace.
\newblock \emph{arXiv preprint arXiv:1812.04754}.

\bibitem[{Hendrycks and Dietterich(2018)}]{hendrycks2018benchmarking}
Hendrycks, D.; and Dietterich, T. 2018.
\newblock Benchmarking Neural Network Robustness to Common Corruptions and Perturbations.
\newblock In \emph{International Conference on Learning Representations}.

\bibitem[{Hendrycks and Dietterich(2019)}]{hendrycks2019benchmarking}
Hendrycks, D.; and Dietterich, T. 2019.
\newblock Benchmarking neural network robustness to common corruptions and perturbations.
\newblock In \emph{International Conference on Learning Representations}.

\bibitem[{Hu et~al.(2024)Hu, Zhang, Sun, Chen, Kuo, and Nevatia}]{hu2024bafta}
Hu, X.; Zhang, K.; Sun, M.; Chen, A.; Kuo, C.-H.; and Nevatia, R. 2024.
\newblock BaFTA: Backprop-Free Test-Time Adaptation For Zero-Shot Vision-Language Models.
\newblock \emph{arXiv preprint arXiv:2406.11309}.

\bibitem[{Iwasawa and Matsuo(2021)}]{iwasawa2021test}
Iwasawa, Y.; and Matsuo, Y. 2021.
\newblock Test-time classifier adjustment module for model-agnostic domain generalization.
\newblock \emph{Advances in Neural Information Processing Systems}, 34: 2427--2440.

\bibitem[{Jang, Chung, and Chung(2023)}]{jang2023test}
Jang, M.; Chung, S.-Y.; and Chung, H.~W. 2023.
\newblock Test-Time Adaptation via Self-Training with Nearest Neighbor Information.
\newblock In \emph{The Twelfth International Conference on Learning Representations}.

\bibitem[{Ji and Telgarsky(2020)}]{NEURIPS2020_c76e4b2f}
Ji, Z.; and Telgarsky, M. 2020.
\newblock Directional convergence and alignment in deep learning.
\newblock In Larochelle, H.; Ranzato, M.; Hadsell, R.; Balcan, M.; and Lin, H., eds., \emph{Advances in Neural Information Processing Systems}, volume~33, 17176--17186. Curran Associates, Inc.

\bibitem[{Kim et~al.(2023)Kim, Sun, Raghunathan, and Kolter}]{kim2023reliable}
Kim, E.; Sun, M.; Raghunathan, A.; and Kolter, Z. 2023.
\newblock Reliable Test-Time Adaptation via Agreement-on-the-Line.
\newblock \emph{arXiv preprint arXiv:2310.04941}.

\bibitem[{Kirichenko, Izmailov, and Wilson(2023)}]{kirichenko2023last}
Kirichenko, P.; Izmailov, P.; and Wilson, A.~G. 2023.
\newblock Last Layer Re-Training is Sufficient for Robustness to Spurious Correlations.
\newblock In \emph{The Eleventh International Conference on Learning Representations}.

\bibitem[{Krizhevsky, Hinton et~al.(2009)}]{krizhevsky2009learning}
Krizhevsky, A.; Hinton, G.; et~al. 2009.
\newblock Learning multiple layers of features from tiny images.
\newblock \emph{Toronto, ON, Canada}.

\bibitem[{Lee et~al.(2024)Lee, Bai, Pres, Wattenberg, Kummerfeld, and Mihalcea}]{lee2024a}
Lee, A.; Bai, X.; Pres, I.; Wattenberg, M.; Kummerfeld, J.~K.; and Mihalcea, R. 2024.
\newblock A Mechanistic Understanding of Alignment Algorithms: A Case Study on {DPO} and Toxicity.
\newblock In \emph{Forty-first International Conference on Machine Learning}.

\bibitem[{Lee et~al.(2013)}]{lee2013pseudo}
Lee, D.-H.; et~al. 2013.
\newblock Pseudo-label: The simple and efficient semi-supervised learning method for deep neural networks.
\newblock In \emph{Workshop on challenges in representation learning, ICML}, volume~3, 896. Atlanta.

\bibitem[{Lee et~al.(2023)Lee, Chen, Tajwar, Kumar, Yao, Liang, and Finn}]{lee2023surgical}
Lee, Y.; Chen, A.~S.; Tajwar, F.; Kumar, A.; Yao, H.; Liang, P.; and Finn, C. 2023.
\newblock Surgical Fine-Tuning Improves Adaptation to Distribution Shifts.
\newblock In \emph{The Eleventh International Conference on Learning Representations}.

\bibitem[{Li et~al.(2017)Li, Yang, Song, and Hospedales}]{li2017deeper}
Li, D.; Yang, Y.; Song, Y.-Z.; and Hospedales, T.~M. 2017.
\newblock Deeper, broader and artier domain generalization.
\newblock In \emph{Proceedings of the IEEE international conference on computer vision}, 5542--5550.

\bibitem[{Li et~al.(2021)Li, Tan, Tao, Liu, and Huang}]{li2021low}
Li, T.; Tan, L.; Tao, Q.; Liu, Y.; and Huang, X. 2021.
\newblock Low dimensional landscape hypothesis is true: DNNs can be trained in tiny subspaces.
\newblock \emph{arXiv preprint arXiv:2103.11154}.

\bibitem[{Liang, He, and Tan(2023)}]{liang2023comprehensive}
Liang, J.; He, R.; and Tan, T.-P. 2023.
\newblock A Comprehensive Survey on Test-Time Adaptation under Distribution Shifts.
\newblock \emph{International Journal of Computer Vision}.

\bibitem[{Liang, Hu, and Feng(2020)}]{liang2020we}
Liang, J.; Hu, D.; and Feng, J. 2020.
\newblock Do we really need to access the source data? source hypothesis transfer for unsupervised domain adaptation.
\newblock In \emph{International conference on machine learning}, 6028--6039. PMLR.

\bibitem[{Lin, Roy, and Li(2021)}]{lin2021mood}
Lin, Z.; Roy, S.~D.; and Li, Y. 2021.
\newblock Mood: Multi-level out-of-distribution detection.
\newblock In \emph{Proceedings of the IEEE/CVF conference on Computer Vision and Pattern Recognition}, 15313--15323.

\bibitem[{Liu et~al.(2020)Liu, Bai, Jiang, Chen, and Wang}]{Liu2020Understanding}
Liu, J.; Bai, Y.; Jiang, G.; Chen, T.; and Wang, H. 2020.
\newblock Understanding Why Neural Networks Generalize Well Through GSNR of Parameters.
\newblock In \emph{International Conference on Learning Representations}.

\bibitem[{Mahsereci et~al.(2017)Mahsereci, Balles, Lassner, and Hennig}]{mahsereci2017early}
Mahsereci, M.; Balles, L.; Lassner, C.; and Hennig, P. 2017.
\newblock Early stopping without a validation set.
\newblock \emph{arXiv preprint arXiv:1703.09580}.

\bibitem[{Michalkiewicz et~al.(2023)Michalkiewicz, Faraki, Yu, Chandraker, and Baktashmotlagh}]{Michalkiewicz_2023_ICCV}
Michalkiewicz, M.; Faraki, M.; Yu, X.; Chandraker, M.; and Baktashmotlagh, M. 2023.
\newblock Domain Generalization Guided by Gradient Signal to Noise Ratio of Parameters.
\newblock In \emph{Proceedings of the IEEE/CVF International Conference on Computer Vision (ICCV)}, 6177--6188.

\bibitem[{Murali et~al.(2023)Murali, Puli, Yu, Ranganath, and Batmanghelich}]{murali2023beyond}
Murali, N.; Puli, A.; Yu, K.; Ranganath, R.; and Batmanghelich, K. 2023.
\newblock Beyond Distribution Shift: Spurious Features Through the Lens of Training Dynamics.
\newblock \emph{Transactions on machine learning research}, 2023.

\bibitem[{Niu et~al.(2022)Niu, Wu, Zhang, Chen, Zheng, Zhao, and Tan}]{niu2022efficient}
Niu, S.; Wu, J.; Zhang, Y.; Chen, Y.; Zheng, S.; Zhao, P.; and Tan, M. 2022.
\newblock Efficient test-time model adaptation without forgetting.
\newblock In \emph{International conference on machine learning}, 16888--16905. PMLR.

\bibitem[{Niu et~al.(2023)Niu, Wu, Zhang, Wen, Chen, Zhao, and Tan}]{niu2023towards}
Niu, S.; Wu, J.; Zhang, Y.; Wen, Z.; Chen, Y.; Zhao, P.; and Tan, M. 2023.
\newblock Towards Stable Test-time Adaptation in Dynamic Wild World.
\newblock In \emph{The Eleventh International Conference on Learning Representations}.

\bibitem[{Panigrahi et~al.(2023)Panigrahi, Saunshi, Zhao, and Arora}]{panigrahi2023task}
Panigrahi, A.; Saunshi, N.; Zhao, H.; and Arora, S. 2023.
\newblock Task-Specific Skill Localization in Fine-tuned Language Models.
\newblock In \emph{International conference on machine learning}. PMLR.

\bibitem[{Parascandolo et~al.(2021)Parascandolo, Neitz, Orvieto, Gresele, and Sch{\"{o}}lkopf}]{ParascandoloNOG21}
Parascandolo, G.; Neitz, A.; Orvieto, A.; Gresele, L.; and Sch{\"{o}}lkopf, B. 2021.
\newblock Learning explanations that are hard to vary.
\newblock In \emph{9th International Conference on Learning Representations, {ICLR}}.

\bibitem[{Park et~al.(2024)Park, Kim, Kwon, Yoon, and Sohn}]{park2024layer}
Park, J.; Kim, J.; Kwon, H.; Yoon, I.; and Sohn, K. 2024.
\newblock Layer-wise Auto-Weighting for Non-Stationary Test-Time Adaptation.
\newblock In \emph{Proceedings of the IEEE/CVF Winter Conference on Applications of Computer Vision}, 1414--1423.

\bibitem[{Pasad, Shi, and Livescu(2023)}]{pasad2023comparative}
Pasad, A.; Shi, B.; and Livescu, K. 2023.
\newblock Comparative layer-wise analysis of self-supervised speech models.
\newblock In \emph{ICASSP 2023-2023 IEEE International Conference on Acoustics, Speech and Signal Processing (ICASSP)}, 1--5. IEEE.

\bibitem[{Rajasegaran et~al.(2019)Rajasegaran, Hayat, Khan, Khan, and Shao}]{rajasegaran2019random}
Rajasegaran, J.; Hayat, M.; Khan, S.~H.; Khan, F.~S.; and Shao, L. 2019.
\newblock Random path selection for continual learning.
\newblock \emph{Advances in neural information processing systems}, 32.

\bibitem[{Russell et~al.(2008)Russell, Torralba, Murphy, and Freeman}]{russell2008labelme}
Russell, B.~C.; Torralba, A.; Murphy, K.~P.; and Freeman, W.~T. 2008.
\newblock LabelMe: a database and web-based tool for image annotation.
\newblock \emph{International journal of computer vision}, 77: 157--173.

\bibitem[{Sankararaman et~al.(2020)Sankararaman, De, Xu, Huang, and Goldstein}]{sankararaman2020impact}
Sankararaman, K.~A.; De, S.; Xu, Z.; Huang, W.~R.; and Goldstein, T. 2020.
\newblock The impact of neural network overparameterization on gradient confusion and stochastic gradient descent.
\newblock In \emph{International conference on machine learning}, 8469--8479. PMLR.

\bibitem[{Shi et~al.(2022)Shi, Seely, Torr, Narayanaswamy, Hannun, Usunier, and Synnaeve}]{ShiSTNHUS22}
Shi, Y.; Seely, J.; Torr, P. H.~S.; Narayanaswamy, S.; Hannun, A.~Y.; Usunier, N.; and Synnaeve, G. 2022.
\newblock Gradient Matching for Domain Generalization.
\newblock In \emph{The Tenth International Conference on Learning Representations, {ICLR}}.

\bibitem[{Shin et~al.(2024)Shin, Lee, Lee, Park, Lee, Hwang, and Yoon}]{shin2024gradient}
Shin, J.; Lee, J.; Lee, S.; Park, M.; Lee, D.; Hwang, U.; and Yoon, S. 2024.
\newblock Gradient Alignment with Prototype Feature for Fully Test-time Adaptation.
\newblock \emph{arXiv preprint arXiv:2402.09004}.

\bibitem[{Sorrenti et~al.(2023)Sorrenti, Bellitto, Salanitri, Pennisi, Spampinato, and Palazzo}]{sorrenti2023selective}
Sorrenti, A.; Bellitto, G.; Salanitri, F.~P.; Pennisi, M.; Spampinato, C.; and Palazzo, S. 2023.
\newblock Selective Freezing for Efficient Continual Learning.
\newblock In \emph{Proceedings of the IEEE/CVF International Conference on Computer Vision}, 3550--3559.

\bibitem[{Su, Xu, and Jia(2022)}]{su2022revisiting}
Su, Y.; Xu, X.; and Jia, K. 2022.
\newblock Revisiting realistic test-time training: Sequential inference and adaptation by anchored clustering.
\newblock \emph{Advances in Neural Information Processing Systems}, 35: 17543--17555.

\bibitem[{Sun et~al.(2020{\natexlab{a}})Sun, Panda, Feris, and Saenko}]{sun2020adashare}
Sun, X.; Panda, R.; Feris, R.; and Saenko, K. 2020{\natexlab{a}}.
\newblock Adashare: Learning what to share for efficient deep multi-task learning.
\newblock \emph{Advances in Neural Information Processing Systems}, 33: 8728--8740.

\bibitem[{Sun et~al.(2020{\natexlab{b}})Sun, Wang, Liu, Miller, Efros, and Hardt}]{sun2020test}
Sun, Y.; Wang, X.; Liu, Z.; Miller, J.; Efros, A.; and Hardt, M. 2020{\natexlab{b}}.
\newblock Test-time training with self-supervision for generalization under distribution shifts.
\newblock In \emph{International conference on machine learning}, 9229--9248. PMLR.

\bibitem[{Suteu and Guo(2019)}]{suteu2019regularizing}
Suteu, M.; and Guo, Y. 2019.
\newblock Regularizing deep multi-task networks using orthogonal gradients.
\newblock \emph{arXiv preprint arXiv:1912.06844}.

\bibitem[{Venkateswara et~al.(2017)Venkateswara, Eusebio, Chakraborty, and Panchanathan}]{venkateswara2017deep}
Venkateswara, H.; Eusebio, J.; Chakraborty, S.; and Panchanathan, S. 2017.
\newblock Deep hashing network for unsupervised domain adaptation.
\newblock In \emph{Proceedings of the IEEE conference on computer vision and pattern recognition}, 5018--5027.

\bibitem[{Vianna et~al.(2023)Vianna, Chaudhary, Tang, Cloutier, Wolf, Eickenberg, and Belilovsky}]{vianna2023channel}
Vianna, P.; Chaudhary, M.~S.; Tang, A.; Cloutier, G.; Wolf, G.; Eickenberg, M.; and Belilovsky, E. 2023.
\newblock Channel Selection for Test-Time Adaptation Under Distribution Shift.
\newblock \emph{NeurIPS 2023 Workshop on Distribution Shifts: New Frontiers with Foundation Models}.

\bibitem[{Wallingford et~al.(2022)Wallingford, Li, Achille, Ravichandran, Fowlkes, Bhotika, and Soatto}]{wallingford2022task}
Wallingford, M.; Li, H.; Achille, A.; Ravichandran, A.; Fowlkes, C.; Bhotika, R.; and Soatto, S. 2022.
\newblock Task adaptive parameter sharing for multi-task learning.
\newblock In \emph{Proceedings of the IEEE/CVF Conference on Computer Vision and Pattern Recognition}, 7561--7570.

\bibitem[{Wang et~al.(2020)Wang, Shelhamer, Liu, Olshausen, and Darrell}]{wang2020tent}
Wang, D.; Shelhamer, E.; Liu, S.; Olshausen, B.; and Darrell, T. 2020.
\newblock Tent: Fully test-time adaptation by entropy minimization.
\newblock \emph{arXiv preprint arXiv:2006.10726}.

\bibitem[{Wang et~al.(2022)Wang, Fink, Van~Gool, and Dai}]{wang2022continual}
Wang, Q.; Fink, O.; Van~Gool, L.; and Dai, D. 2022.
\newblock Continual test-time domain adaptation.
\newblock In \emph{Proceedings of the IEEE/CVF Conference on Computer Vision and Pattern Recognition}, 7201--7211.

\bibitem[{Wang et~al.(2023)Wang, Zhang, Yan, Zhang, and Li}]{wang2023feature}
Wang, S.; Zhang, D.; Yan, Z.; Zhang, J.; and Li, R. 2023.
\newblock Feature alignment and uniformity for test time adaptation.
\newblock In \emph{Proceedings of the IEEE/CVF Conference on Computer Vision and Pattern Recognition}, 20050--20060.

\bibitem[{Wang et~al.(2024)Wang, Luo, Zheng, Chen, Wang, and Huang}]{wang2024search}
Wang, Z.; Luo, Y.; Zheng, L.; Chen, Z.; Wang, S.; and Huang, Z. 2024.
\newblock In Search of Lost Online Test-time Adaptation: A Survey.
\newblock \emph{International Journal of Computer Vision (IJCV)}.

\bibitem[{Wortsman et~al.(2022)Wortsman, Ilharco, Gadre, Roelofs, Gontijo-Lopes, Morcos, Namkoong, Farhadi, Carmon, Kornblith et~al.}]{wortsman2022model}
Wortsman, M.; Ilharco, G.; Gadre, S.~Y.; Roelofs, R.; Gontijo-Lopes, R.; Morcos, A.~S.; Namkoong, H.; Farhadi, A.; Carmon, Y.; Kornblith, S.; et~al. 2022.
\newblock Model soups: averaging weights of multiple fine-tuned models improves accuracy without increasing inference time.
\newblock In \emph{International Conference on Machine Learning}, 23965--23998. PMLR.

\bibitem[{Wu et~al.(2024{\natexlab{a}})Wu, Wang, Huang, Zheng, Zhao, and Wei}]{wu2024enhanced}
Wu, Y.; Wang, H.; Huang, L.-K.; Zheng, Y.; Zhao, P.; and Wei, Y. 2024{\natexlab{a}}.
\newblock Enhanced Gradient Aligned Continual Learning via Pareto Optimization.
\newblock In \emph{International Conference on Learning Representations}.

\bibitem[{Wu et~al.(2024{\natexlab{b}})Wu, Wang, Zhao, Zheng, Wei, and Huang}]{wumitigating}
Wu, Y.; Wang, H.; Zhao, P.; Zheng, Y.; Wei, Y.; and Huang, L.-K. 2024{\natexlab{b}}.
\newblock Mitigating Catastrophic Forgetting in Online Continual Learning by Modeling Previous Task Interrelations via Pareto Optimization.
\newblock In \emph{Forty-first International Conference on Machine Learning}.

\bibitem[{Xie et~al.(2017)Xie, Girshick, Doll{\'a}r, Tu, and He}]{xie2017aggregated}
Xie, S.; Girshick, R.; Doll{\'a}r, P.; Tu, Z.; and He, K. 2017.
\newblock Aggregated residual transformations for deep neural networks.
\newblock In \emph{Proceedings of the IEEE conference on computer vision and pattern recognition}, 1492--1500.

\bibitem[{Yaras et~al.(2023)Yaras, Wang, Hu, Zhu, Balzano, and Qu}]{yaras2023law}
Yaras, C.; Wang, P.; Hu, W.; Zhu, Z.; Balzano, L.; and Qu, Q. 2023.
\newblock The law of parsimony in gradient descent for learning deep linear networks.
\newblock \emph{arXiv preprint arXiv:2306.01154}.

\bibitem[{Yu et~al.(2020)Yu, Kumar, Gupta, Levine, Hausman, and Finn}]{yu2020gradient}
Yu, T.; Kumar, S.; Gupta, A.; Levine, S.; Hausman, K.; and Finn, C. 2020.
\newblock Gradient surgery for multi-task learning.
\newblock \emph{Advances in Neural Information Processing Systems}, 33: 5824--5836.

\bibitem[{Yu et~al.(2023)Yu, Shin, Lee, Jun, and Lee}]{yu2023block}
Yu, Y.; Shin, S.; Lee, S.; Jun, C.; and Lee, K. 2023.
\newblock Block selection method for using feature norm in out-of-distribution detection.
\newblock In \emph{Proceedings of the IEEE/CVF Conference on Computer Vision and Pattern Recognition}, 15701--15711.

\bibitem[{Yuan, Xie, and Li(2023)}]{yuan2023robust}
Yuan, L.; Xie, B.; and Li, S. 2023.
\newblock Robust test-time adaptation in dynamic scenarios.
\newblock In \emph{Proceedings of the IEEE/CVF Conference on Computer Vision and Pattern Recognition}, 15922--15932.

\bibitem[{Yuan, Feng, and Liu(2024)}]{yuan2024early}
Yuan, S.; Feng, L.; and Liu, T. 2024.
\newblock Early Stopping Against Label Noise Without Validation Data.
\newblock In \emph{International Conference on Learning Representations}.

\bibitem[{Yuan et~al.(2024)Yuan, He, Dong, Han, and Yin}]{yuan2024discriminability}
Yuan, Y.; He, R.; Dong, Y.; Han, Z.; and Yin, Y. 2024.
\newblock Discriminability-Driven Channel Selection for Out-of-Distribution Detection.
\newblock In \emph{Proceedings of the IEEE/CVF Conference on Computer Vision and Pattern Recognition}, 26171--26180.

\bibitem[{Zagoruyko and Komodakis(2016)}]{zagoruyko2016wide}
Zagoruyko, S.; and Komodakis, N. 2016.
\newblock Wide Residual Networks.
\newblock In \emph{British Machine Vision Conference}. British Machine Vision Association.

\bibitem[{Zhang, Levine, and Finn(2022)}]{zhang2022memo}
Zhang, M.; Levine, S.; and Finn, C. 2022.
\newblock Memo: Test time robustness via adaptation and augmentation.
\newblock \emph{Advances in Neural Information Processing Systems}, 35: 38629--38642.

\bibitem[{Zhang et~al.(2024)Zhang, Wan, Zhang, Cheung, Tian, Shen, and Ye}]{zhang2024interpreting}
Zhang, W.; Wan, C.; Zhang, Y.; Cheung, Y.-m.; Tian, X.; Shen, X.; and Ye, J. 2024.
\newblock Interpreting and Improving Large Language Models in Arithmetic Calculation.
\newblock In \emph{Forty-first International Conference on Machine Learning}.

\bibitem[{Zhang et~al.(2022)Zhang, Chen, Cheng, Li, Li, Lin, and Li}]{zhang2022divide}
Zhang, Z.; Chen, W.; Cheng, H.; Li, Z.; Li, S.; Lin, L.; and Li, G. 2022.
\newblock Divide and contrast: Source-free domain adaptation via adaptive contrastive learning.
\newblock \emph{Advances in Neural Information Processing Systems}, 35: 5137--5149.

\bibitem[{Zhao, Chen, and Xia(2023)}]{zhao2023delta}
Zhao, B.; Chen, C.; and Xia, S.-T. 2023.
\newblock DELTA: DEGRADATION-FREE FULLY TEST-TIME ADAPTATION.
\newblock In \emph{The Eleventh International Conference on Learning Representations}.

\bibitem[{Zhao et~al.(2023{\natexlab{a}})Zhao, Liu, Alahi, and Lin}]{zhao2023pitfalls}
Zhao, H.; Liu, Y.; Alahi, A.; and Lin, T. 2023{\natexlab{a}}.
\newblock On Pitfalls of Test-Time Adaptation.
\newblock In \emph{International conference on machine learning}. PMLR.

\bibitem[{Zhao et~al.(2023{\natexlab{b}})Zhao, Zhou, Long, Jiang, and Zhang}]{zhao2023does}
Zhao, H.; Zhou, T.; Long, G.; Jiang, J.; and Zhang, C. 2023{\natexlab{b}}.
\newblock Does continual learning equally forget all parameters?
\newblock In \emph{International Conference on Machine Learning}, 42280--42303. PMLR.

\end{thebibliography}

\clearpage
\appendix

\twocolumn[
  \centering
  \section*{\normalfont\LARGE\bfseries Supplementary Materials for \\``A Layer Selection Approach to Test Time Adaptation''} 
  \vspace{2em} 
]

\renewcommand{\thesection}{A.\arabic{section}}
\setcounter{figure}{5}
\setcounter{table}{4}
\setcounter{equation}{8}

In the supplementary section, we provide a comprehensive discussion of the experimental setups and the proposed approach, \LEAST{}. The supplementary material is organized as follows:

\begin{itemize}
    \item \textbf{Sec. \ref{supp:sec:extendedRelatedWork}}: Related works section.
    
    \item \textbf{Sec. \ref{supp:sec:domainbed}}: Implementation details of the Domainbed benchmark (for Tables \ref{tab:dg_bench_results}, \ref{tab:analysis}, and Figure \ref{fig:heatmap}).

    \item \textbf{Sec. \ref{supp:sec:continualTTA}}: Implementation details of the Continual TTA benchmark (for Tables \ref{tab:cotta} and \ref{tab:analysis}).

    \item \textbf{Sec. \ref{supp:sec:tinydomainbed}}: Implementation details of the Tiny Domainbed benchmark (for Table \ref{tab:study}), including the rationale for selecting critical shifts from the Domainbed benchmark.

    \item \textbf{Sec. \ref{supp:sec:LEASTdiscussion}}: An in-depth discussion of \LEAST{}, including pseudocode and an analysis of its balance between gradient alignment and magnitude.

    \item \textbf{Sec. \ref{supp:sec:additionalResults}}: Additional experimental result tables and plots not included in the main paper.

    \item \textbf{Sec. \ref{disc}}: Discussion section.

\end{itemize}

The numbering of figures, tables, and equations in this supplementary material continues from the main paper to ensure consistency and avoid repetition.

\section{Related Works}
\label{supp:sec:extendedRelatedWork}

\subsection{Regularization in test time adaptation}
Various regularization approaches have been proposed to address the problem of degradation in TTA. While few works try to improve the quality of pseudolabels used for adaptation \cite{chen2022contrastive, wang2022continual}, other set of works try to improve class prototypes \cite{iwasawa2021test, jang2023test, hu2024bafta}. Constraining the model by aligning source and target domain features has also been explored in various works \cite{su2022revisiting, zhang2022divide, gao2023back}. Also, a lot of works have proposed different regularization terms to the TTA formulation to prevent degradation \cite{niu2022efficient, zhang2022memo, zhao2023delta, niu2023towards, shin2024gradient} However, in this paper, we take a parameter-centric approach to regularize test time adaptation, which we argue can be more effective for efficient TTA. Moreover, these regularization approaches can be used in conjunction with our proposed approaches to improve TTA performance further.

\subsection{Layer selection}
Layer selection or identifying the optimal set of parameters for a certain task has been important for a number of fields. Layer selection can help improve the performance in fine-tuning. Layer selection can improve sharing of features across different tasks which has been found to be beneficial in multi-task learning \cite{ahn2019deep, guo2020learning, wallingford2022task, sun2020adashare},
and continual learning \cite{rajasegaran2019random, sorrenti2023selective, zhao2023does}. While good layer selection can help us better understand and explain features learnt by the model \cite{lee2024a, zhang2024interpreting}, identifying the right layer to take features from can be important for out-of-distribution detection \cite{lin2021mood, yu2023block, elaraby2023grood, darrin2024unsupervised, yuan2024discriminability}. In pretraining, layer selection can help create robust pretrained models for domain generalization \cite{2020EccvDMG}, or can result in different amounts of performance gains in
self supervised learning \cite{gidaris2018unsupervised, bordes2023guillotine, pasad2023comparative}. Closer to our field, identifying the right set of parameters to train has shown to have big impacts in performance in 
fine-tuning \cite{guo2019spottune, wortsman2022model, lee2023surgical, panigrahi2023task} and learning robust non-spurious features \cite{kirichenko2023last, murali2023beyond}. There have been few works \citet{ vianna2023channel, lee2023surgical, park2024layer} that study the differential impact of parameters on TTA but the impact of these studies on parameter selection is somewhat limited to individual TTA losses, rendering them non-exhaustive and potentially inapplicable to other TTA losses proposed in the past or future. On the other hand, we propose a gradient aligned layer adaptation framework for TTA, which is more flexible than existing layer selection strategies, and we demonstrate it improves performance across different TTA losses.

\subsection{Gradient alignment studies}
\citet{gur2018gradient} discuss the phenomenon of gradient descent occurring within a tiny subspace. \citet{yaras2023law} demonstrate the presence of a low-dimensional structure in learning dynamics. \citet{li2021low} reveal that neural networks can be effectively trained in lower-dimensional subspaces. \citet{gauch2022few} show that gradient descent within a tiny subspace enhances generalization in few-shot learning. \citet{Liu2020Understanding} indicate that a high gradient signal-to-noise ratio can lead to improved generalization in neural networks. \citet{NEURIPS2020_c76e4b2f} find that the gradients of neural networks converge to a single direction. \citet{sankararaman2020impact} show that gradient alignment accelerates the training speed of neural networks. Building upon these approaches, this paper proposes a novel cosine distance-based criterion for layer selection in the context of test-time adaptation.\\

\subsection{Applications of gradient alignment}
\citet{andriushchenko2020understanding} and \citet{gao2021gradient} propose using gradient alignment as a regularizer for adversarial training. \citet{barba2021implicit} apply gradient alignment within the context of federated learning. \citet{fort2019stiffness} demonstrate that gradient alignment can be useful for detecting overfitting. \citet{gupta2020look}, along with \citet{wumitigating, wu2024enhanced}, show that gradient alignment can mitigate catastrophic forgetting in continual learning. \citet{Michalkiewicz_2023_ICCV}, \citet{ParascandoloNOG21} and \citet{ShiSTNHUS22} utilize gradient alignment for domain generalization. \citet{yu2020gradient} show that gradient alignment aids in multi-task learning, a finding supported by \citet{du2018adapting} and \citet{suteu2019regularizing}. Building on these approaches, we propose a novel formulation of gradient alignment for an online and unsupervised application of test time adaptation. \\

\subsection{Gradient alignment-based early stopping}
Recent works have proposed various approaches or criteria for early stopping without a validation set. \citet{mahsereci2017early} introduced an evidence-based criterion based on the variance of gradients \cite{agarwal2022estimating}. \citet{forouzesh2021disparity} proposed using gradient disparity across samples as a criterion. \citet{yuan2024early} suggested performing early stopping by tracking the model’s predictions on the samples. Most of these existing works have demonstrated the effectiveness of their proposed approaches, often in the context of multi-epoch and supervised learning \cite{bonet2021channel} or noisy learning \cite{bai2021understanding}. Similar to these approaches, our novel cosine distance-based criterion can be used to perform layer-wise early stopping without a validation set, preventing degradation in test-time adaptation and beyond.

\section{Experimental Details of Domainbed}
\label{supp:sec:domainbed}

\subsection{Dataset Details}
\label{supp:subsec:domainbed:benchmark}
Domainbed \cite{gulrajani2021search} consists of four domain generalization datasets: 
\begin{itemize}
    \item PACS \cite{li2017deeper} consists of different object images from four domains: Art, Cartoon, Photo, and Sketch. It comprises of 9,991 samples across 7 class labels (i.e., dog, elephant, giraffe, guitar, horse, house, and person).

    \item VLCS \cite{fang2013unbiased} consists of photographic images from four domains/datasets: PASCAL VOC207 \cite{everingham2010pascal}, LabelMe \cite{russell2008labelme}, Caltech 101 \cite{fei2004learning}, and SUN09\cite{choi2010exploiting}. It comprises of 10,729 samples across 5 class labels (i.e., bird, car, chair, dog, and person).

    \item TerraIncognita \cite{beery2018recognition} consists of images of wild animals taken at different locations, which make up the four domains: L100, L38, L43, and L46. It comprises of 24,788 samples across 10 class labels (i.e., bird, bobcat, cat, coyote, dog, empty/no animal, opossum, rabbit, raccoon, squirrel).

    \item Office-Home \cite{venkateswara2017deep} consists of different object images typically seen in offices and homes from four domains: Art, Clipart, Product, and Real World. It comprises of 15,588 samples across 65 class labels (e.g., bottle, computer, hammer, pen).
\end{itemize}

\subsection{Evaluation Details}
\label{supp:subsec:domainbed:evaluation}
\begin{figure*}[t!]
  \centering
  \begin{subfigure}{0.45\linewidth}
    \centering
    \includegraphics[scale=0.325]{images/res18_PL_new.png}
    \caption{Resnet-18 with PL loss}
    \label{supp:fig:res18_PL}
  \end{subfigure}
  \begin{subfigure}{0.45\linewidth}
    \centering
    \includegraphics[scale=0.325]{images/res18_SHOT_new.png}
    \caption{Resnet-18 with SHOT loss}
    \label{supp:fig:res18_SHOT}
  \end{subfigure}
  \par\bigskip
  \begin{subfigure}{0.45\linewidth}
    \centering
    \includegraphics[scale=0.325]{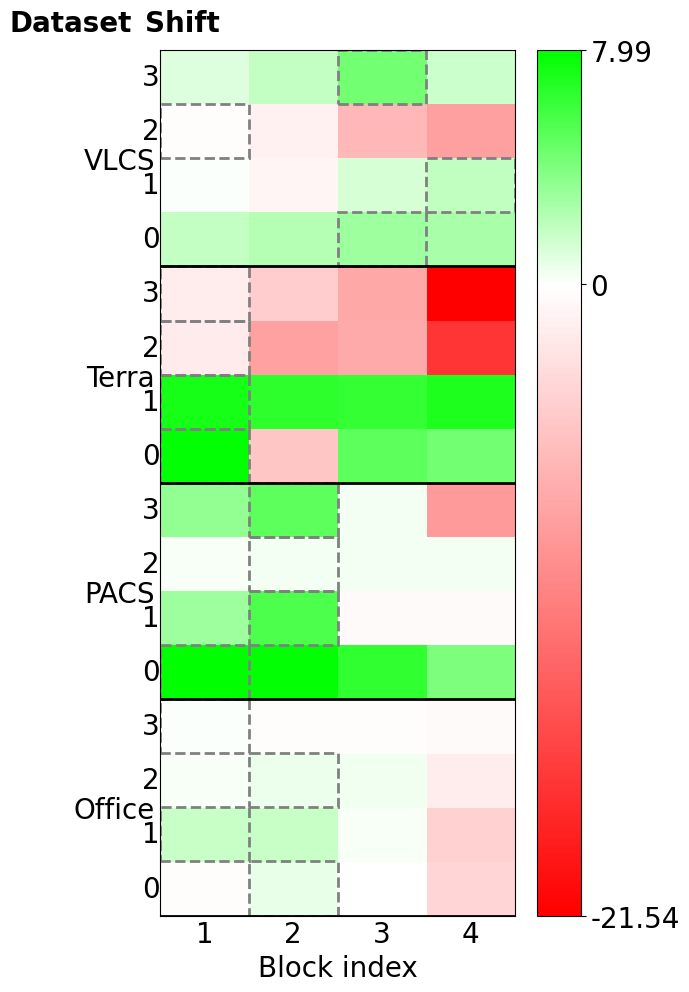}
    \caption{Resnet-50 with PL loss}
    \label{supp:fig:res50_PL}
  \end{subfigure}
  \begin{subfigure}{0.45\linewidth}
    \centering
    \includegraphics[scale=0.325]{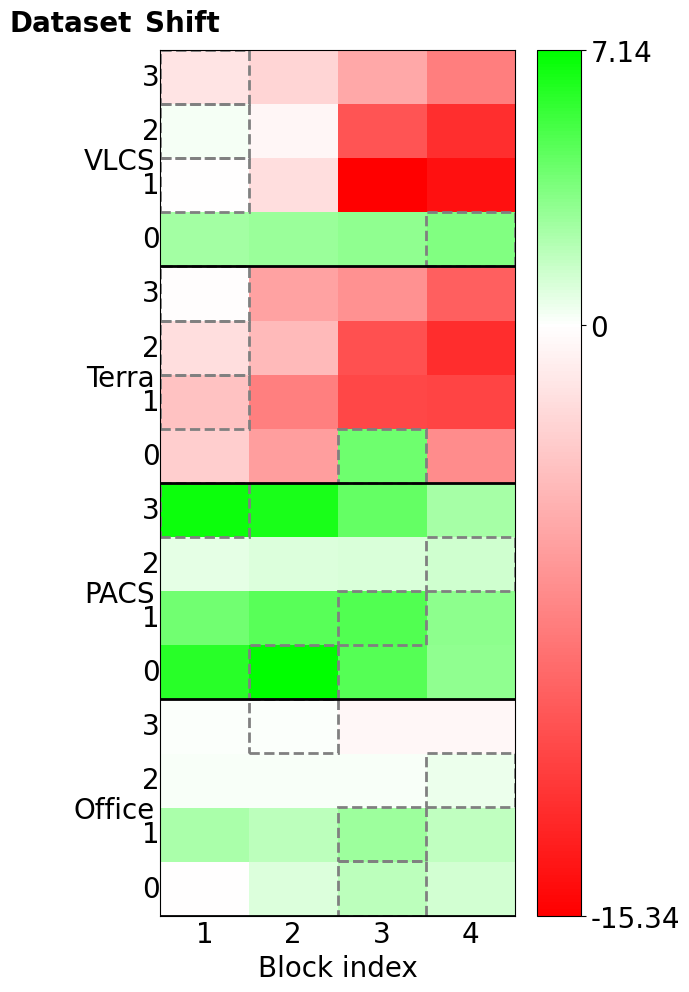}
    \caption{Resnet-50 with SHOT loss}
    \label{supp:fig:res50_SHOT}
  \end{subfigure}
  \caption{Heatmap of Performance improvement (\%) per-block on Domainbed benchmark for Resnet-18 (same as Figure 4) and Resnet-50. Performance improvement is the difference between the TTA accuracy of a given block/layer and ERM accuracy for the same shift. Positive performance improvements are shown in green, and negative performance improvements (or degradation) are in red. Using the bounding box, we highlight the best block per loss and dataset shift.}
  \label{supp:fig:heatmap}
\end{figure*}

We follow the evaluation protocol as described in \citet{iwasawa2021test}. In Domainbed, the pretrained model is trained on all but one domain. All the domains on which the pretrained model is trained are referred to as training domains, and the remaining domain is referred to as the testing domain. We follow the dataset splits used in T3A \cite{iwasawa2021test}. Each domain is split into a big and a small split. Specifically, the domains are split into 80\% and 20\%. The big split of training domains is used for training the pretrained model and is referred to as training splits. The small splits of training domains are referred to as validation splits. The big split of the testing domain is used to evaluate the domain and is referred to as the testing split. This is the split where test time adaptation is performed for each minibatch before inference. We consider three seeds for the results in Tab. \ref{tab:dg_bench_results} and one seed in Tab. \ref{tab:analysis}, Fig. \ref{fig:heatmap}, and Fig. \ref{supp:fig:heatmap}. Each seed generates a new training and testing split from training and testing domains. See Figure 1 ``Data configuration for a benchmark with four domains'' in Domainbed \cite{gulrajani2021search} supplementary.

The performance on each domain or shift is obtained by averaging across multiple seeds. Next, we obtain the performance on each dataset by averaging across all domains in the dataset. Finally, we get performance on Domainbed by averaging across all the datasets in Domainbed.

\subsection{Hyperparameters and Model Selection}
\label{supp:subsec:domainbed:hp}
\subsubsection{Pretrained Model}
We follow the pretraining protocol as described in \citet{iwasawa2021test}. We use ERM for pretraining the model similar to other works in TTA \cite{iwasawa2021test, jang2023test, wang2023feature}. We consider two backbones: Resnet-18 and Resnet-50, with batch normalization layers. The backbone networks are trained using ERM and Adam optimizer with a batch size of 32. We follow the training-domain validation-based model selection \cite{gulrajani2021search, iwasawa2021test} where we choose the hyperparameters that maximize the accuracy of the pretrained model on the validation splits. Please refer to Domainbed \cite{gulrajani2021search} and T3A \cite{iwasawa2021test} for a detailed discussion on hyperparameters and the range used.

\subsubsection{TTA Approaches}
We follow the adaptation protocol as described in \citet{iwasawa2021test}. Similar to other works in TTA \cite{zhao2023pitfalls, kim2023reliable, boudiaf2022parameter, gong2022note}, our hyperparameter tuning protocol of TTA methods is based on \citet{zhao2023pitfalls}, where we choose the best hyperparameter set for each TTA method under consideration. We consider two popular TTA methods from \citet{iwasawa2021test}: pseudo-labeling (PL) \cite{lee2013pseudo} and SHOT \cite{liang2020we}.
Please refer to \citet{iwasawa2021test} for a detailed discussion on hyperparameters and the range used.

\subsubsection{Baseline Layer Selection Methods}
We perform a model selection for each baseline as described in the original implementations. ERM \cite{gulrajani2021search, iwasawa2021test}, \textit{All Layers} \cite{iwasawa2021test}, and AutoRGN \cite{lee2023surgical} baselines do not have any hyperparameters. The hyperparameter for AutoSNR \cite{lee2023surgical} baseline is tuned as described in \citet{lee2023surgical}.


\section{Experimental Details of Continual TTA}
\label{supp:sec:continualTTA}

\subsection{Dataset Details}

Continual TTA benchmark \cite{wang2022continual} consists of two datasets, CIFAR10-C and CIFAR100-C, widely used for evaluating the robustness of classification networks under various corruptions, particularly in the context of test-time adaptation (TTA). These datasets are derived from the original CIFAR10 and CIFAR100 \cite{krizhevsky2009learning}, which contain 50,000 training and 10,000 test images across 10 and 100 categories, respectively. In CIFAR10-C and CIFAR100-C \cite{hendrycks2018benchmarking}, 15 types of corruptions, each with 5 levels of severity, are applied to the test images of their clean counterparts. This results in 10,000 corrupted images for each corruption type in both the datasets.

\begin{algorithm}[t]
\caption{Gradient-Aligned Layer Adaptation (GALA)}
\label{supp:algo:LEAST}
\begin{algorithmic}[1]
\STATE \textbf{Initialization:} Pretrained model: $f_{\mathbf{\theta}_0}(x)$, Anchor model parameters: $\mathbf{\theta}_{\text{anchor}} = \mathbf{\theta}_0$, Adaptation method: \texttt{TTA}, Window size: $s$, Mask threshold: $\lambda$

\STATE \textbf{Input for step $i$:} Sample $x_i$, anchor $\mathbf{\theta}_{\text{anchor}}$ and current model $\theta_{i-1}$

    \STATE $\mathbf{u}_{i} = \texttt{TTA}(x_i, \mathbf{\theta}_{i-1})$    
    \FOR{each layer $l$}
        \STATE $\mathbf{TD}_{i-1,l} = \mathbf{\theta}_{i-1,l} - \mathbf{\theta}_{\text{anchor},l}$
        \STATE $\mathrm{cos}(\alpha_{i,l}) =\frac{\mathbf{u}_{i,l} \cdot (\mathbf{u}_{i,l}+\mathbf{TD}_{i-1,l})}{\Vert\mathbf{u}_{i,l}\Vert_2 ~\Vert\mathbf{u}_{i,l}+\mathbf{TD}_{i-1,l}\Vert_2}$ 
        \STATE $m_{i,l} = 
            \begin{cases} 
            1 & \text{if $\mathrm{cos}(\alpha_{i,l}) > \lambda$} \\
            0 & \text{otherwise}
            \end{cases}$
        \STATE $\mathbf{\theta}_{i,l} = \mathbf{\theta}_{i-1,l} + m_{i,l} \mathbf{u}_{i,l}$  
    \ENDFOR
    \STATE $r = \lfloor (i-1) / s \rfloor$
    \IF{$i == r*s$} 
        \STATE $\mathbf{\theta}_{\text{anchor}} = \mathbf{\theta}_i$
    \ENDIF
\STATE \textbf{Output at step $i$:} Prediction $f_{\mathbf{\theta}_i}(x_i)$, Updated model $f_{\mathbf{\theta}_i}$
\end{algorithmic}
\end{algorithm}

\subsection{Evaluation Details}
We follow the evaluation protocol as described in \citet{wang2022continual}. We utilize a model pre-trained on the clean training set of the CIFAR10 or CIFAR100 dataset. During test time, corrupted images are provided to the network in an online fashion. We continually adapt the source pretrained model to each corruption type sequentially without resetting to the pretrained model. The CIFAR10 and CIFAR100 experiments follow this online continual test-time adaptation scheme, with evaluations conducted under the highest corruption severity level 5. The evaluation is based on the online prediction results immediately after encountering the data.

\subsection{Hyperparameters and Model Selection}
\noindent \textbf{Pretrained Model} \quad
We follow the pretraining protocol as described in \citet{wang2022continual}. For our experiments on CIFAR10C and CIFAR100C, we utilize pre-trained models from the RobustBench benchmark \cite{croce2robustbench} similar to previous works in test time adaptation \cite{wang2020tent, wang2022continual, niu2022efficient, yuan2023robust}. Specifically, for CIFAR10C, we employ a WideResNet-28 \cite{zagoruyko2016wide} model, and for CIFAR100C, we adopt a pre-trained ResNeXt-29 \cite{xie2017aggregated} model, which is one of the default architectures for CIFAR100 in RobustBench. 

\noindent \textbf{TTA Approaches} \quad
We follow the evaluation protocol as described in \citet{wang2022continual}.  We update the model with one gradient step per test point at each iteration, utilizing the Adam optimizer with a learning rate of 1e-3. The hyperparameters employed are consistent with those recommended by \citet{wang2022continual}. To facilitate comparison with Domainbed benchmark results, we incorporate the same two test-time adaptation methods: pseudo-labeling (PL) \cite{lee2013pseudo} and SHOT \cite{liang2020we}.

\noindent \textbf{Baseline Layer Selection Methods} \quad
We perform a model selection for each baseline as described in the original implementations. ERM \cite{croce2robustbench, wang2022continual} and \textit{All Layers} \cite{iwasawa2021test} baselines do not have any hyperparameters.

\section{Experimental Details of Tiny Domainbed}
\label{supp:sec:tinydomainbed}
\subsection{Dataset and Shift Details}
\label{supp:subsec:tinydomainbed:benchmark}
We create Tiny-Domainbed from Domainbed by selecting the following critical shifts: 
\begin{itemize}
\item Three shifts from the Terra Incognita benchmark: L100, L38, and  L43;
\item Two shifts from the PACS benchmark: Cartoon and Sketch;
\item  One shift from the VLCS benchmark: SUN09.
\end{itemize}

\subsection{Discussion on Chosen Shifts}
\label{supp:subsec:tinyDomainbed:chosenshifts}

Creating Tiny-Domainbed aims to make the smallest possible setup of Domainbed, which contains all the challenging shifts or domains in Domainbed while being computationally light for ease of analysis and comparison. To identify the critical shifts of Domainbed, we refer to the heatmap of performance improvement of blocks vs. shifts in Domainbed in Fig. \ref{supp:fig:heatmap}. We refer to blocks and layers interchangeably in this section.

Based on heatmaps of the Resnet-18 backbone for pseudolabelling and SHOT loss-based TTA methods in Fig. \ref{supp:fig:heatmap}, we identify the critical shifts in Domainbed which satisfy the following two important properties: 

\begin{figure*}[t]
    \centering
    \includegraphics[width=0.8\linewidth]{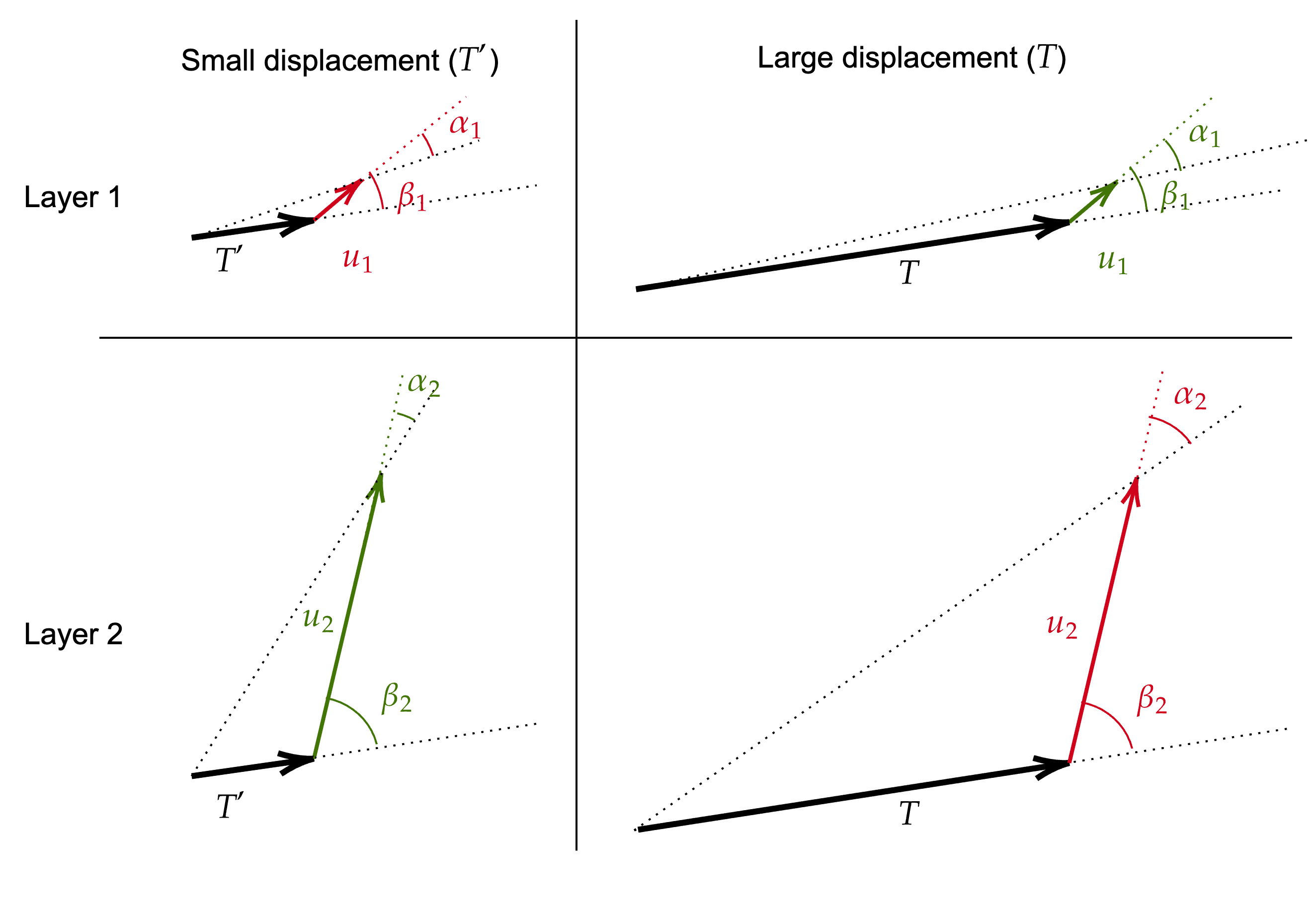}
    \caption{Effect of magnitude of $\mathbf{u}$ on cosine distance criterion. Consider two vectors such that $u_1$ is smaller than $u_2$ but is better aligned with its displacement. The rows illustrate the layers, and the columns denote the two scenarios. The vectors are shown in green if the cosine distance selects the layer, or else shown in red. \textbf{Left}: In scenario 2 of small displacements ($T^{'}$), the update’s magnitude can dominate the criterion, and \LEAST{} selects $u_1$. \textbf{Right} In scenario 1 of large displacements ($T$), alignment becomes crucial, and \LEAST{} selects $u_2$.}
    \label{supp:fig:magn2_fourplots}
\end{figure*}

\begin{itemize}
    \item \textbf{Property 1: Shifts with brightest green/red blocks.} \quad
    A bright green layer implies that adapting this layer improves performance over ERM. Similarly, a bright red layer implies that adapting this layer can degrade the performance with respect to ERM. The shifts with the brightest green or red layers are important because any layer selection criterion must do well on these shifts. The inability to choose bright green layers while adapting to these shifts is a missed opportunity for performance improvement of the layer selection approach. Again, being unable to avoid bright red layers in these shifts can result in significant performance degradation. In Fig. \ref{supp:fig:heatmap}, we see that the following shifts on the Resnet-18 backbone have the brightest green/red layers: $env0, env1, env3$ in PACS; $env2, env3$ in VLCS; $env0, env1, env2, env3$ in Terra Incognita; \textit{none} in OfficeHome.

    \item \textbf{Property 2: Shifts whose best block changes with the TTA loss function.} \quad
    We make a striking observation that for certain shifts in Domainbed, the best block for a given shift can depend on the TTA loss used. This implies that the layer selection criterion must consider TTA loss to choose the best layers to adapt. In Fig. \ref{supp:fig:heatmap}, we see that the following shifts on the Resnet-18 backbone experience a change in the location of the best block due to a change in the TTA method: $env1$ in PACS; $env0, env3$ in VLCS; $env0$ in Terra Incognita; $env1, env2$ in OfficeHome.
    
\end{itemize}

To create the Tiny-Domainbed benchmark, we select the most critical shifts with the brightest green/red blocks, and the location of the best block changes with the TTA loss function. Based on this, we identify the following shifts from Domainbed to be included in the Tiny-Domainbed benchmark: 

\begin{itemize}
    \item Shifts that satisfy both the properties: $env3$ in PACS; $env3$ in VLCS; $env0$ in Terra Incognita.

    \item Shifts that only satisfy property 1 but are included in Tiny Domainbed: $env1$ in PACS; $env1, env2$ in Terra Incognita.
\end{itemize}

This gives us our final list of critical shifts from Domainbed included in the Tiny-Domainbed benchmark: three shifts from the Terra Incognita benchmark: L100, L38, and  L43; two shifts from the PACS benchmark: Cartoon and Sketch, and one shift from the VLCS benchmark: SUN09.

\subsection{Evaluation Details}
\label{supp:subsec:tinydomainbed:evaluation}
We follow the evaluation protocol as described in \citet{iwasawa2021test} and is described in detail in Sec. \ref{supp:subsec:domainbed:evaluation}. We obtain the performance on a given testing domain similar to the evaluation protocol of Domainbed. However, we obtain the final performance on Tiny-Domainbed by averaging across only the selected domains or shifts (identified as critical shifts in Sec. \ref{supp:subsec:tinydomainbed:benchmark}) on the Resnet-18 backbone and two TTA losses.

\noindent \textbf{Evaluation Metrics}. \quad
We will explain the various metrics employed to compare different block or layer selection methods, as used in Table \ref{tab:study}:
\begin{itemize}
    \item \textbf{TTA Accuracy}: We abbreviate it as \textit{TTA acc}. It refers to the accuracy of testing samples from the target domain observed during adaptation. This metric follows the same evaluation protocol as in Tables \ref{tab:dg_bench_results} and \ref{tab:cotta} to generate the TTA results, providing insight into the performance of different layer selection approaches at test time.

    \item \textbf{Generalization}: It measures the accuracy on the held-out split of the target domain after the model has completed adaptation on all the samples from the same target domain. This metric indicates how each layer selection method aids in adapting the model to the target domain.

    \item \textbf{Forgetting}: This metric quantifies the drop in accuracy on the held-out split of source domains after the pretrained model has adapted to all samples of the target domain, which differs from the source domains on which the pretrained model was initially trained. This metric helps us assess the degree of forgetting of source features due to various layer selection methods.

    \item \textbf{Rank Correlation}: In this metric, we measure the Spearman correlation of layer selection ranks between the oracle and proposed layer selection methods. Oracle TTA performance, as depicted in Figure \ref{supp:fig:heatmap}, ranks the four blocks for each configuration. Similarly, different layer selection methods adapt a layer with a specific frequency during TTA, resulting in a ranking of the four blocks. We evaluate the relationship between these two ranking methods using Spearman rank correlation. This correlation provides insight into how well the proposed layer selection methods’ ranking of layers aligns with the oracle ranking on Tiny-Domainbed.

\end{itemize}

\begin{table}[t]
\centering
\resizebox{0.9\linewidth}{!}{%
\begin{tabular}{lc|cc|cc}
\toprule
\multicolumn{1}{l}{\textbf{}}                                & \multicolumn{1}{l}{} & \multicolumn{2}{c}{\textbf{Resnet-18} $\uparrow$}           & \multicolumn{2}{c}{\textbf{Resnet-50}$\uparrow$}                               \\
\multicolumn{1}{l}{\textbf{Setting}}                         & \multicolumn{1}{c}{\textbf{Condition}}   & \multicolumn{1}{c}{\textbf{PL}} & \multicolumn{1}{c}{\textbf{SHOT}} & \multicolumn{1}{c}{\textbf{PL}} & \multicolumn{1}{c}{\textbf{SHOT}} \\
\midrule
\multirow{3}{*}{\textbf{Partitioning}}                       & Single block         & $66.19$       & $65.8$                              & $70.21$                           & $68.37$                             \\
                                                             & Single layer         & $66.31$       & $66.78$                             & $71.07$                           & $70.13$                             \\
                                                             & Multiple layers      & $63.13$       & $65.36$                             & $69.11$                           & $68.29$                             \\
\midrule
\multirow{3}{*}{\textbf{Threshold}}                          & $0.5$                  & $66.31$       & $66.78$                             & $71.07$                           & $70.13$                             \\
                                                             & $0.75$                 & $66.31$       & $66.78$                             & $71.07$                           & $70.13$                             \\
                                                             & $0.99$                 & $66.24$       & $66.96$                             & $71.07$                           & $70.13$                             \\
\midrule
\multirow{3}{*}{\textbf{Window Size}}                        & $5$                   & $66.37$       & $66.61$                             & $70.8$7                           & $70.01$                             \\
                                                             & $20$                   & $66.31$       & $66.78$                             & $71.07$                           & $70.13$                             \\
                                                             & $\infty$                    & $65.71$       & $66.59$                             & $70.76$                           & $70.42$                             \\
\midrule
\multicolumn{1}{l}{\multirow{2}{*}{\textbf{Batch Size = 1}}} & All Layers           & $37.47$       & $29.67$                             & $33.93$                           & $32.81$                             \\
\multicolumn{1}{l}{}                                         & GALA                 & $64.12$       & $64.29$                             & $70.45$                           & $70.24$                            \\
\bottomrule
\end{tabular}%
}
\caption{Accuracy (\%) under different experimental conditions. The values are averaged for each backbone and TTA loss of the Domainbed benchmark.}
\label{supp:tab:analysis2_fullDB}
\end{table}

\subsection{Hyperparameters and Model Selection}
\label{supp:subsec:tinydomainbed:hp}
We follow the model selection and hyperparameter tuning protocol for pretrained models and TTA approaches as described in \citet{iwasawa2021test} and is described in detail Sec. \ref{supp:subsec:domainbed:hp}. We consider only the Resnet-18 backbone with the batch normalization layers for ease of analysis. We consider two TTA losses: pseudolabelling and SHOT. We use \textit{Block} granularity-based layer selection and report results for a single seed for easy analysis. Please note that although we perform block-based layer selection in Tiny-Domainbed, we interchangeably refer to block selection or layer selection in this section.

\noindent \textbf{Layer Selection Methods}. \quad
In Table \ref{tab:study}, we compare the \LEAST{} method with the following oracle (\textit{Best Block} and \textit{Worst Block}) and baseline methods (All Blocks and Random Block):
\begin{itemize}
    \item \textbf{All Blocks}: This is analogous to the \textit{All Layers} baseline in Tables \ref{tab:dg_bench_results} and \ref{tab:cotta}, where all blocks of the model are adapted at each adaptation step. 
    
    \item \textbf{Random Block}: During each adaptation step for a given sample, a block is chosen at random and adapted accordingly.
    
    \item \textbf{Best Block}: In this oracle layer selection method, the best-performing block for each shift and loss function, as identified in Figure \ref{supp:fig:heatmap}, is adapted for all adaptation steps of the model.
    
    \item \textbf{Worst Block}: This oracle method adapts the worst-performing block for each shift and loss function, as identified in Figure \ref{supp:fig:heatmap}, for all adaptation steps of the model.
\end{itemize}

\section{Discussion on \LEAST{}}
\label{supp:sec:LEASTdiscussion}

\begin{table}[t]
\centering
\resizebox{0.85\linewidth}{!}{%
\begin{tabular}{lc|cc|cc}
\toprule
\textbf{}                                                   & \multicolumn{1}{l}{} & \multicolumn{2}{c}{\textbf{CIFAR10C} $\uparrow$}                               & \multicolumn{2}{c}{\textbf{CIFAR100C} $\uparrow$}                              \\
\textbf{Setting}                                            & \multicolumn{1}{c}{\textbf{Condition}}   & \multicolumn{1}{c}{\textbf{PL}} & \multicolumn{1}{c}{\textbf{SHOT}} & \multicolumn{1}{c}{\textbf{PL}} & \multicolumn{1}{c}{\textbf{SHOT}} \\
\midrule
\multicolumn{1}{c}{\multirow{2}{*}{\textbf{Continual TTA}}} & No Reset             & $73.6$                            & $76.41$                             & $64.75$                           & $64.87$                             \\
\multicolumn{1}{c}{}                                        & With Reset           & $71.32$                           & $79.54$                             & $66.31$                           & $67.13$                            \\
\bottomrule
\end{tabular}%
}
\caption{Accuracy (\%) under different experimental conditions. The values are averaged for each dataset and TTA loss of the Continual TTA benchmark.}
\label{supp:tab:analysis2_fullCoTTA}
\end{table}

\subsection{Pseudocode}
A pseudocode of \LEAST{} is given in Algorithm \ref{supp:algo:LEAST}.

\subsection{Implementation details of \LEAST{}}
\LEAST{} has two hyperparameters, namely \textit{window size} and \textit{mask threshold}. In Sec. \ref{sec:analysisLEAST}, we show that \LEAST{} is not overly sensitive to its hyperparameters and, therefore, use a fixed value (\textit{window size} = 20 and \textit{mask threshold}=0.75) across all setups when comparing with the baselines. Total displacement computed over the initial few samples may be unreliable, which can result in incorrect layers selected for adaptation. We address this by scaling the masked updates for an initial few samples in the reset window. One could also use an earlier anchor model from the previous reset window, but scaling the mask for a few initial minibatches seems to suffice.

Based on Sec. \ref{sec:analysisLEAST}, we note that \LEAST{} performs the best when the pretrained model is adapted only with the best single layer identified by \LEAST{} (the one with the largest cosine distance above the selection threshold) but not with multiple or the top $k$-best layers identified by \LEAST{} (i.e., layers whose cosine distance is larger than the selection threshold). Therefore, we use \LEAST{} with Single-layer partitioning across all setups when comparing with the baselines. An important point to note is that \LEAST{} adapts all the layers for the first sample in a reset window. It adapts the most gradient-aligned layer per sample for all the other samples. (However, in the Tiny-Domainbed benchmark, we use Single-block partitioning since the analysis is performed at Block granularity.)

\begin{figure*}[t]
    \centering
    \begin{subfigure}{0.3\linewidth}
        \centering
        \includegraphics[scale=0.4]{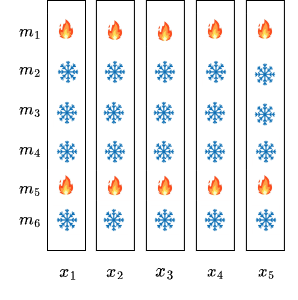}
        \caption{Layer-wise}
    \end{subfigure}
    \begin{subfigure}{0.3\linewidth}
        \centering
        \includegraphics[scale=0.4]{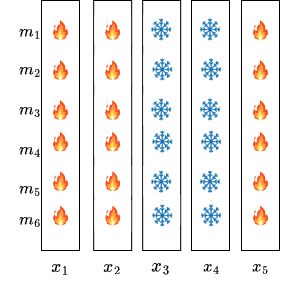}
        \caption{Sample-wise}
    \end{subfigure}
    \begin{subfigure}{0.3\linewidth}
        \centering
        \includegraphics[scale=0.4]{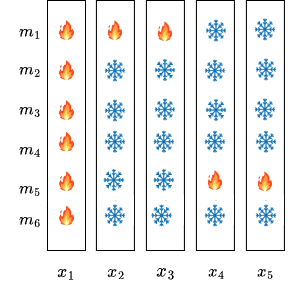}
        \caption{\LEAST{}}
    \end{subfigure}
    \caption{Different adaptation strategies: (a) TTA approaches typically adapt a fixed set of layers for all the samples. (b) Sample selection-based TTA approaches skip the adaptation of all layers on a few unreliable samples. (c) \LEAST{} is more flexible and can dynamically control the adaptation of individual layers per sample.}
    \label{supp:fig:ls_vs_es_les}
\end{figure*}

Since \LEAST{} performs the best with Single-layer partitioning instead of Multi-layer partitioning, it implies that \LEAST{} can often identify a \good{} layer to adapt but may not identify all the top $k$-best layers to adapt above the selection threshold. This can be viewed as one of the limitations of \LEAST{} and can potentially be a fruitful research direction for future works. We note that one can address this limitation by tuning the hyperparameters of \LEAST{} (especially \textit{mask threshold}) in each setup, similar to the recommendation made by \citet{zhao2023pitfalls}. However, we avoid any hyperparameter tuning of \LEAST{} in the paper and show that adapting the model with the best single-layer identified by \LEAST{} can outperform existing baselines.

\subsection{Relationship between \LEAST{} and Eq. \ref{eq:cosAlphaBeta_new}}
For notational simplicity, we rewrite the following terms

\begin{itemize}
    \item $\mathbf{u} = \mathbf{u}_{i,l}$, the current sample's update for layer $l$.

    \item $u = \Vert\mathbf{u}\Vert_2$, the magnitude of the current sample's update.
    
    \item $\mathbf{T} = \mathbf{TD}_{i-1,l}$, the total displacement undergone in previous steps by layer $l$.

    \item $T = \Vert\mathbf{T}\Vert_2$, the magnitude of the total displacement. This is the \textit{magnitude} used in Sec. \ref{sec:analysisLEAST} and Sec. \ref{supp:subsec:align_vs_magn}.

    \item $\beta = \beta_{i,l}$ is the angle between $\mathbf{u}$ and $\mathbf{T}$.

    \item $\cos(\beta)$ is the alignment of $\mathbf{u}$ with $\mathbf{T}$. This is the \textit{alignment} used in Sec. \ref{sec:analysisLEAST} and Sec. \ref{supp:subsec:align_vs_magn}. We also interchangeably refer to it as \textit{direction} since $\beta$ is the angle $\mathbf{u}$ makes with $\mathbf{T}$.

    \item $\alpha = \alpha_{i,l}$ is the angle between $\mathbf{u}$ and $\mathbf{u} + \mathbf{T}$.

    \item $\cos(\alpha)$ is the proposed criterion of \LEAST{}. In Sec. \ref{sec:analysisLEAST} and Sec. \ref{supp:subsec:align_vs_magn}, we note that the proposed cosine distance criterion effectively balances \textit{magnitude} and \textit{alignment}.
\end{itemize}

Based on the above definitions, the \textit{alignment} is given by

\begin{equation}    \label{supp:eq:cosBeta}
\cos(\beta) = \frac{\mathbf{u} \cdot \mathbf{T}}{u~T}
\end{equation}

Let us begin by expanding the numerator of Eq. \ref{eq:cosine},

\begin{align}
\cos(\alpha) & = \frac{\mathbf{u} \cdot (\mathbf{T}+\mathbf{u})}{u\Vert\mathbf{T}+\mathbf{u} \Vert_2 } \\
     & = \frac{\mathbf{u} \cdot \mathbf{T} + \mathbf{u} \cdot \mathbf{u}}{u\Vert\mathbf{T}+\mathbf{u} \Vert_2 } \\
      & = \frac{uT \cos(\beta) + u^2}{u \Vert\mathbf{T}+\mathbf{u} \Vert_2 }  \\
     & = \frac{T\cos(\beta)+u}{\Vert\mathbf{T}+\mathbf{u} \Vert_2}
\end{align}

If $\beta$ is acute, then we can see that using the Pythagorean theorem, we get $\Vert\mathbf{T}+\mathbf{u} \Vert_2 = \sqrt{(T+u \cos(\beta))^2+(u\sin(\beta))^2}$. One can show that this also holds for obtuse $\beta$, in which case $u \cos(\beta)$ is negative. Substituting this in the equation above, we get our Eq. \ref{eq:cosAlphaBeta_new} as 

\begin{align}   \label{supp:eq:cosAlphaBeta_new}
\cos(\alpha) & = \frac{T \cos(\beta) + u}{\sqrt{(T+u \cos(\beta))^2+(u\sin(\beta))^2}}.
\end{align}

\subsection{Alignment vs Magnitude in \LEAST{}}
\label{supp:subsec:align_vs_magn}
In this section, we expand on the discussion of Sec. \ref{sec:analysisLEAST} to better understand how the proposed cosine distance criterion depends on the \textit{magnitude} and the \textit{alignment} of the current sample's update with respect to the total displacement made so far.

From Eq.~\ref{supp:eq:cosAlphaBeta_new}, it is clear that computing the proposed cosine distance criterion for a given layer only involves $T$, $u$, and the angle $\beta$. This means that even though the layers may have a considerable number of parameters, we can always draw a diagram like the ones in Fig.~\ref{supp:fig:magn2_fourplots} to represent the situation and compare the updates for performing layer selection. 

To better understand the interaction between magnitude and alignment towards cosine metric, we consider an example where we have two layers, layer 1 and layer 2, and their corresponding total displacement has the same norm $T$. We consider an update $\textbf{u}_1$ for layer 1  and an update $\textbf{u}_2$ for layer 2 with: $u_1 < u_2$. While the update for layer 2 has a larger magnitude, the update $\textbf{u}_1$ of layer 1 is more aligned with its displacement ($\beta_1 < \beta_2$). We observe that two scenarios can arise depending on the magnitude of the total displacement:

\begin{itemize}
    \item \textbf{Scenario 1 of large $T$}: In this scenario, Cosine distance significantly depends on the \textit{alignment} of the current sample and is less impacted by its \textit{magnitude}. This scenario is more likely for most of the samples during TTA. This can also be viewed similarly to performing an exploit strategy in a given update direction, i.e., after having seen a certain number of samples, we allow adaptation of a layer if the gradients for new updates are well aligned with the previously seen updates.
    
    \item \textbf{Scenario 2 of small $T$}: In this scenario, Cosine distance significantly depends on the \textit{magnitude} of the current sample and is less impacted by its \textit{alignment} with the total displacement. This can occur for the very first samples in a few cases or if the gradients on a layer for a particular sample are much larger than usual. Similarly, we can view it as performing a form of pure exploration strategy over the update directions.
\end{itemize}

Fig. \ref{fig:align_vs_magn} (left) and Fig. \ref{supp:fig:magn2_fourplots} visualize the two scenarios for the above example.

\section{Additional Experimental Results}
\label{supp:sec:additionalResults}

In this section, we include the following additional experimental results not included in the main paper:
\begin{itemize}
    \item \textbf{Performance improvement heatmap for Resnet-50 backbone on Domainbed benchmark}: While Fig. \ref{fig:heatmap} shows the heatmap for performance improvement only for the Resnet-18 backbone, in Fig. \ref{supp:fig:heatmap}, we also show the heatmap for performance improvement for the Resnet-50 backbone for comparison. Similar to the heatmap for Resnet-18, we observe that no single layer of Resnet-50 is suitable for all settings, and not all layers are equally receptive to adaptation. Moreover, similar to the Resnet-18's backbone, we observe that the location of \good{} layers can change across the shifts of a given dataset and for a TTA loss function even for the same shift.

    \item \textbf{Effect of experimental conditions on different backbones and TTA loss functions on Domainbed benchmark}: While the first four settings of Table \ref{tab:analysis} show the effect of experimental conditions averaged on the whole of Domainbed, Tab. \ref{supp:tab:analysis2_fullDB} shows a more fine-grained impact of experimental conditions by reporting the values averaged on each backbone and TTA loss function of the Domainbed. Similar to the discussion in Sec. \ref{sec:analysisLEAST}, we observe that Layer granularity performs better than Block granularity, and adaptation with the best Single-layer is much better than with the best Multiple-layers. Tuning the reset window size can improve performance, and the choice of selection threshold is not very sensitive. Finally, \LEAST{} improves over \textit{All Layers} baseline on the single sample adaptation setting (with batch size = 1).

    \item \textbf{Effect of reset on different datasets of Continual TTA benchmark}: While the last setting of Tab. \ref{tab:analysis} shows the effect of reset averaged on the whole of Continual TTA, Tab. \ref{supp:tab:analysis2_fullCoTTA} shows a more fine-grained effect of reset by reporting the values averaged on each dataset and TTA loss function of the Continual TTA. Our observations indicate that the \LEAST{}, when utilizing a reset mechanism with a window size of 20, is advantageous in most scenarios. While it does not appear to benefit the CIFAR10C dataset with PL loss-based TTA, tuning the reset window size could help improve the performance of the reset mechanism.
\end{itemize}

\section{Discussion}
\label{disc}
The simplicity and versatility of \LEAST{} enable seamless integration with existing TTA loss functions, making it a valuable tool for enhancing the adaptability and reliability of deep learning models in real-world applications. Beyond its immediate impact on TTA, our work opens up new avenues for future research in areas where regularization is crucial for learning stability, selective parameter updates could be beneficial, or gradient-aligned feature learning might offer additional advantages. \LEAST{} not only advances the field of TTA but also contributes to the broader goal of developing more robust and adaptive AI systems.

\end{document}